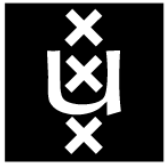

UNIVERSITEIT VAN AMSTERDAM

MSc ARTIFICIAL INTELLIGENCE
MASTER THESIS

---

## Hyperbolic Convolutional Neural Networks

---

by
ANDRII SKLIAR
11636785

August 27, 2019

36 European Credits
February 2019 - August 2019

*Supervisor:*  *Assessor:*
MSc. Maurice Weiler  Dr. Efstratios Gavves

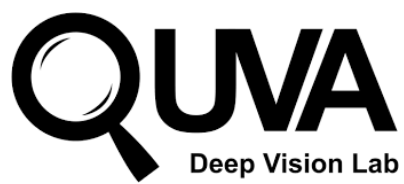

# Abstract

Deep Learning is mostly responsible for the surge of interest in Artificial Intelligence in the last decade. So far, deep learning researchers have been particularly successful in the domain of image processing, where Convolutional Neural Networks are used. Although excelling at image classification, Convolutional Neural Networks are quite naive in that no inductive bias is set on the embedding space for images. Similar flaws are also exhibited by another type of Convolutional Networks - Graph Convolutional Neural Networks. However, using non-Euclidean space for embedding data might result in more robust and explainable models. One example of such a non-Euclidean space is hyperbolic space. Hyperbolic spaces are particularly useful due to their ability to fit more data in a low-dimensional space and tree-likeliness properties. These attractive properties have been previously used in multiple papers which indicated that they are beneficial for building hierarchical embeddings using shallow models and, recently, using MLPs and RNNs.

However, no papers have yet suggested a general approach to using Hyperbolic Convolutional Neural Networks for structured data processing, although these are the most common examples of data used. Therefore, the goal of this work is to devise a general recipe for building Hyperbolic Convolutional Neural Networks. We hypothesize that ability of hyperbolic space to capture hierarchy in the data would lead to better performance. This ability should be particularly useful in cases where data has a tree-like structure. Since this is the case for many existing datasets (Miller, 1995; Deng et al., 2009; Toutanova et al., 2015), we argue that such a model would be advantageous both in terms of applications and future research prospects.

# Acknowledgement

First of all, I would like to say that this Masters program and this project, in particular, was a great journey that enriched me both intellectually and mentally. In large, this is due to the great people that surrounded me throughout it. I can't be more grateful to my classmates, friends, and teachers who made these two years worth it. Long working days and weekends would not be as rewarding if not for your help that guided me, support that allowed me to push myself forward and friendship that let me get through the hardest of times. Thank you for always being there for me.

This project was a demanding yet rewarding experience which allowed me to learn more about incredibly interesting mathematical topics as well as expand my understanding of Deep Learning. It would not be possible without the help, guidance, and support from my supervisor Maurice Weiler. There are not enough words to express my gratitude for the time and effort he has invested in this project. It is due to his patience and dedication that I got to enjoy this project so much. I believe that I could not wish for a better supervisor and hope to continue working together even after my graduation.

I also owe a significant share of gratitude to people who supported me during this project with their helpful comments and ideas. Special thanks to Pascal Esser, Gabriele Cesa, Gautier Dagan, Gabriele Bani, and Davide Belli for the challenging discussions which made my work much more exciting and satisfying. Of course, thank you all the Masters Room people and Cool People who listened to me at all times - whether I was excited about the new ideas or depressed about bad results. I hope I have found ways to show my appreciation for the time we have spent together.

# Contents

# Introduction <span style="float:right">1</span>

The current success of Deep Learning primarily comes from the super-human performance of neural network models in the context of various tasks including image processing, natural language processing and reinforcement learning. This success has attracted much attention from researchers in the field of Machine Learning, which led to recent advances in the field with hundreds of papers being published daily.

However, in this work, we would like to focus on one particular area of Deep Learning, namely, representation learning. Representation learning is one area of paramount importance to the research community since better feature representation can lead to large improvements in the performance of machine learning models. A simple yet insightful example can be seen in fig. 1.1. By using a function for changing the representation of the data, we can make it linearly separable. Better representation further allows us to use simpler models while getting comparable or better performance.

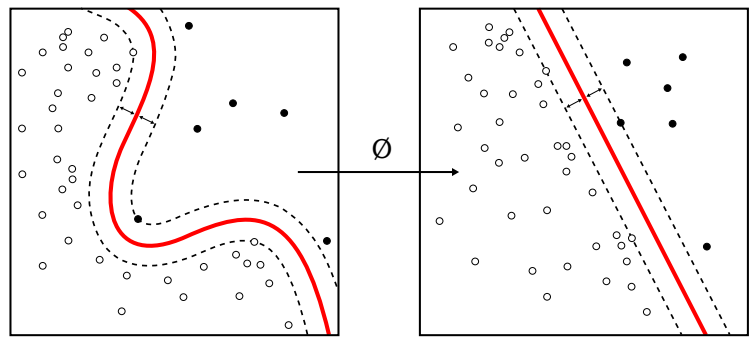

**Fig. 1.1.:** Better representation can lead to improved performance of machine learning models; reproduced from Wikipedia.

However, an important question os how to discover better representations. While in the last decade, handcrafted features were the go-to tool; recently, the focus has shifted to models that would allow us to learn features from the data. Most of these models, albeit powerful, have constraining assumptions on the data embedding space, which, in most cases, is considered to be Euclidean. Such assumptions, however, contradict the well-known manifold hypothesis, which states that "real-world data presented in high dimensional spaces are likely to concentrate in the vicinity of non-linear sub-manifolds of much lower dimensionality" (Rifai et al., 2011).

Different approaches were suggested to tackle this task. They generally belong into two main groups: approaches that would learn data manifold in an unsupervised manner (Posada, 2018; McInnes et al., 2018) and approaches that choose a manifold beforehand and learn an optimal embedding on the specific manifold (Falorsi et al., 2018; Davidson et al., 2018). We argue, however, that these models suffer from the fact that they only use manifolds to embed data as an output of the model. We conjecture that embedding data on a chosen manifold as an intermediate representation at each layer would be highly beneficial for the performance of such models.

Such an approach can be interpreted as imposing an inductive bias on a learning system (Baxter, 2000). Inductive bias leads to an improved generalization since a new learning task is now a relaxed version of the original task with added constraints. We can see changing Euclidean space to hyperbolic space as an example of such inductive bias.

An important point to note is that convolutional neural networks learn to capture more and more complex and abstract features of the provided data with every layer (Zeiler and Fergus, 2014). In the case of images, first layers perform the role of the edge and corner detectors while deeper layers learn to capture class-specific features, e.g., dogs faces, cats paws. Since it is possible to organize classes of images in the tree-like hierarchies (Deng et al., 2009), see chapter 1, we believe that using hyperbolic space would allow us to make use of the hierarchy structure of the classes. We can make use of it by forcing the organization of features in deeper layers to have the same hierarchical structure as classes themselves.

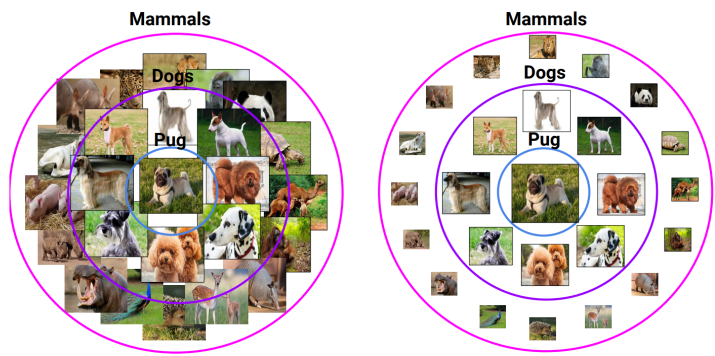

**Fig. 1.2.:** Images can be represented in a hierarchical way. Euclidean space (left) can't account for that unlike hyperbolic space (right); reproduced from Mathieu et al. (2019).

Similarly, we argue that existing graph processing models would benefit from using hyperbolic spaces due to their ability to embed tree-structured data efficiently (Nickel and Kiela, 2017; Ganea et al., 2018b).

To approach these problems, we take inspiration from Ganea et al. (2018a) to build entirely hyperbolic convolutional neural networks. We also suggest Gromov $\delta$-hyperbolicity as a way of measuring whether data can benefit from being embedded in the hyperbolic space.

## 1.1 Research Questions

Throughout this work, we attempt to answer the following research questions:

**RQ1** Can we adapt existing computational blocks, commonly used in deep convolutional neural networks, to embed data in the hyperbolic space?

**RQ2** Can existing models benefit from being fully hyperbolic?

**RQ3** Can we analyze whether data can be effectively embedded into hyperbolic space before running a computationally expensive model?

## 1.2 Contributions

The main contributions of this thesis are:

- Introduction of hyperbolic versions of all operations necessary to implement most of modern deep convolutional neural networks. These are Convolutional Layer, Batch Normalization, Dropout, Pooling, Graph Convolution.

- Application of a theoretical framework of Gromov hyperbolicity as an effective and computationally efficient tool for analyzing whether it is possible to embed data in the hyperbolic space with low distortion.

## 1.3 Outline

The rest of this thesis is structured as follows. In chapter 2 we provide an introduction into main mathematical topics that are necessary for a thorough understanding of this work. Next, in chapter 3, we review recent works that are directly relevant to this thesis. In chapter 4 we introduce hyperbolic operations which will further be used for implementing a fully hyperbolic versions of VGG11 (Simonyan and Zisserman, 2014) and GCN (Kipf and Welling, 2016). Implementation, as well as other details of the experimental setup, are described in chapter 5. Afterwards, in chapter 6, we provide results of experiments and do a thorough discussion of those. Finally, chapter 7 we conclude this work and suggest directions for future research.

# 2 Mathematical Preliminaries

In this chapter, we give a brief introduction of multiple areas of mathematics necessary to understand this work. We first start with the introduction of topological spaces and smooth manifolds as specific cases of those. We then show how, by equipping smooth manifolds with additional structure, Riemannian manifolds arise. Next, using definitions from Riemannian Geometry, we introduce Hyperbolic Spaces as a specific case of Riemannian manifolds. Additionally, we define terms from geometric group theory that provide us with a theoretical framework to work with gyrovector space defined over hyperbolic space.

## 2.1 Topology and Smooth Manifolds

Topology is generally defined as "rubber-sheet geometry" since it is concerned with the study of geometric objects under continuous transformations, i.e., stretching, shrinking, twisting and bending, but not tearing apart or glueing parts together. Importantly, it allows us to define two objects as equivalent if there is such continuous transformation between them, see fig. 2.1. Since we can define two objects as being topologically equivalent, topology also provides us with tools to analyze both objects as representatives of the same equivalence class.

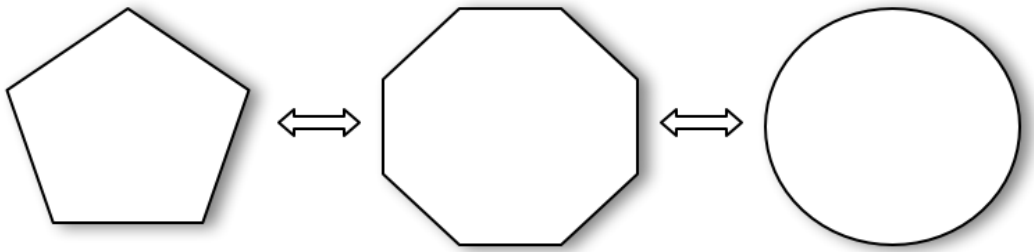

**Fig. 2.1.:** Surfaces of Pentagon, Octagon and Circle can be seed as continuous deformations of each other.

> **Definition 1: Topology (Nakahara, 2003)**
>
> Let $X$ be any set and $\mathscr{T} = \{U_i | i \in I\}$ denote a certain collections of subsets of $X$. The pair $(X, \mathscr{T})$ is a **topological space** if $\mathscr{T}$ satisfies the following requirements:
>
> 1. $\varnothing, X \in \mathscr{T}$
>
> 2. If $J$ is any (maybe infinite) subcollection of $I$, the family $\{U_j | j \in J\}$ satisfies $\bigcup_{j \in J} U_j \in \mathscr{T}$
>
> 3. If $K$ is any *finite* subcollection of $I$, the family $\{U_k | k \in K\}$ satisfies $\bigcap_{k \in K} U_k \in \mathscr{T}$
>
> $X$ alone is sometimes called a topological space. The $U_i$ are called the **open sets** and $\mathscr{T}$ is said to give a **topology** to $X$.

Arguably simplest examples of topologies are:

- Discrete topology: $X$ is an arbitrary set while $\mathscr{T}$ is the collection of all subsets of $X$;
- Trivial topology: $\mathscr{T} = \{\varnothing, X\}$;
- Usual topology: $X = \mathbb{R}$, $\mathscr{T} = \bigcup_{a,b}\{(a,b)\}, a \in \mathbb{R}, b \in \mathbb{R}$.

Another notable example of a topology that we are used to is Euclidean space $\mathbb{R}$. To show that is indeed a topology, we need to define an additional construction, metric, over elements of set $X$. Note, however, that metric is generally not needed for a space to be topological.

> **Definition 2: Metric (Nakahara, 2003)**
>
> A **metric** $d : X \times X \to \mathbb{R}_{>0}$ is a function that for any $x, y, z \in X$ satisfies the conditions:
>
> - Symmetry: $d(x,y) = d(y,x)$
> - Non-negativity: $d(x,y) \geq 0$ with equality holding iff $x = y$
> - Triangle inequality: $d(x,y) + d(y,z) \geq d(x,z)$

Intuitively, a metric can be thought of as distance between objects from the underlying space. In fact, distance in Euclidean space is the most well-known example of a metric.

Metric allows us to define metric topology, where $(X, d)$ is an arbitrary metric space and the topology is induced by the metric. Open sets are defined as a collection of all possible open balls $B(x_0, r) = \{x \in X | d(x_0, x) < r\}$ with $x_o \in X$ and $r > 0$.

By construction of such metric topology, it is easy to see that Euclidean space itself is a topological space. As it was consistent with what humans could observe around themselves, this most straightforward kind of geometry was of central interest to mathematicians for many centuries. However, when it was generally accepted that the Earth is not flat as it was thought before, new rules had to be devised to work with this new, spherical, geometry. Over time this developed into its branch of differential geometry with manifolds lying in the very centre of it.

Informally, a manifold can be defined as a topological space which locally resembles Euclidean space near every point. This is consistent with our view of Earth. Locally, rules from Euclidean geometry are obeyed. However, if we try to apply the same rules to, e.g., airways, we would notice that straight lines are not representing the shortest path anymore. This makes the Earth the simplest yet useful example of a manifold, see fig. 2.2. Note, however, that Euclidean space is also an example of a manifold since it is clearly locally (and also globally) Euclidean.

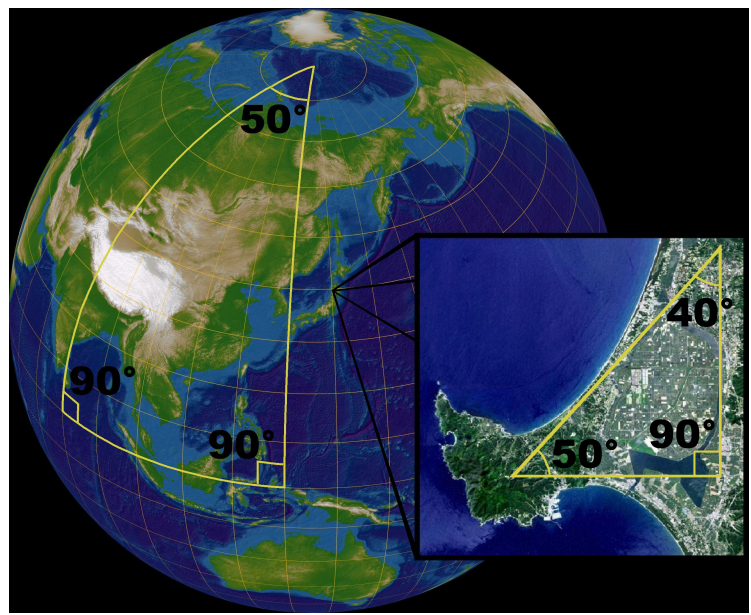

**Fig. 2.2.:** Earth as the simplest example of a manifold; reproduced from Wikipedia.

> **Definition 3: Manifold (Nakahara, 2003)**
>
> $M$ is an $m$-dimensional differentiable **manifold** if
> - $M$ is a topological space
> - $M$ is provided with a family of pairs $\{(U_i, \phi_i)\}$
> - $\{U_i\}$ is a family of open sets which covers $M$, that is, $\bigcup_i U_i = M$. $\phi_i$ is a homeomorphism from $U_i$ onto an open subset $U_i'$ of $\mathbb{R}^m$
> - given $U_i$ and $U_j$ such that $U_i \cap U_j \neq \varnothing$, the map $\phi_{ji} = \phi_i \cdot \phi_j^{-1}$ from $\phi_j(U_i \cap U_j)$ to $\phi_i(U_i \cap U_j)$ is infinitely differentiable i.e. smooth, see fig. 2.3.
>
> A pair $(U_i, \phi_i)$ is called a **chart**. A full set of charts is called an **atlas**. A single $U_i$ is called the **coordinate neighbourhood** while $\phi_i$ is the **coordinate function** or **coordinate**.
>
> In fact, $\phi_i$ can be defined as following: $\phi_i : p \in M \mapsto \{x^1(p), \ldots x^m(p)\}, \forall j \in [1,m] : x^j(p) \in \mathbb{R}$. Then the whole set $\{x^j(p)\}$ can also be called the coordinate. It is important to notice, however, that point $p \in M$ lives on a manifold and, therefore, exists independently from the coordinate system we choose.

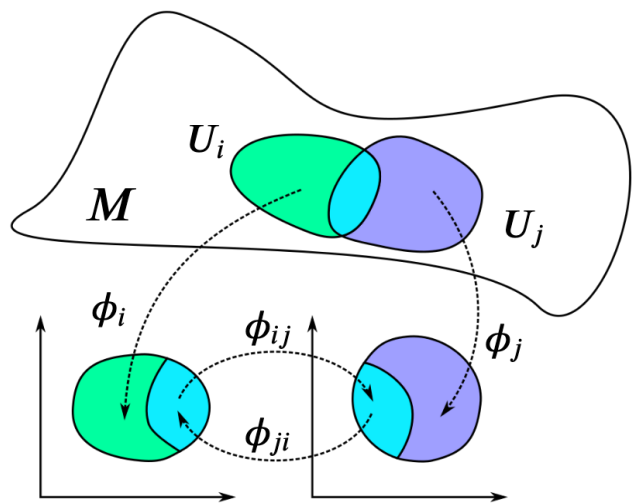

**Fig. 2.3.:** Example of a smooth manifold. Coordinate system $\phi_i$ maps $U_i$ onto an open subset $\phi_i(U_i)$ of $\mathbb{R}$. Since $U_i \cap U_j \neq \varnothing$, $\phi_{ij}$ and $\phi_{ji}$, which represent transitions between two coordinate systems, are infinitely differentiable; reproduced from Wikipedia.

Note that Earth is aligned with our definition of a manifold. Although we can use different coordinate systems to navigate on Earth surface, we will focus on polar coordinates. Polar coordinates $\theta, \phi$ are usually defined as:

$$\theta = tan^{-1}\frac{\sqrt{x^2+y^2}}{z}, \qquad \theta \in [0, \pi]$$
$$\phi = tan^{-1}\frac{y}{x}, \qquad \phi \in [0, 2\pi)$$

Let's take equator $\phi = \frac{1}{2}\pi$ for definiteness. If we try to use it blindly, we will quickly run into a problem of discontinuity on the pole where $\phi$ changes from $2\pi$ to 0. To alleviate this problem, two charts can be defined each one being responsible for a different part of the Earth - one for the south pole and one for the north pole. This is exactly how it was, for example, done in the Universal Polar Stereographic coordinate system.

In the introduction, we have mentioned the manifold hypothesis that states that real-world high-dimensional data lie on low-dimensional manifolds embedded within the high-dimensional space. We use this statement as one of the main motivations of this work. However, the justification for why working on a lower-dimensional manifold is meaningful comes from the following theorem.

> **Theorem 1: Embedding Theorem (Whitney and Eells, 1992)**
>
> Any smooth real $m$-dimensional manifold can be smoothly embedded in the real $2m$-space ($\mathbb{R}^{2m}$), if $m > 0$.

The reason why we have explicitly defined a differentiable manifold and not just an arbitrary topological manifold is that there is no well-defined way to move between charts in a non-differentiable manifold and, therefore, defining any operations over it is problematic. However, the existence of smooth transitions between coordinate systems also allows us to do standard calculus on smooth manifolds.

> **Note**
>
> In the rest of this work, whenever we say manifold, we mean a real $d$-dimensional smooth manifold unless explicitly stated otherwise.

So far, we have defined what a manifold is and have shown how it can be useful. However, there has not been a formalized way of how to work with it yet. To work with manifolds, we need to first define necessary geometric objects, namely, curves and vectors.

> **Definition 4: Smooth Curve**
>
> A **smooth curve** on a manifold $M$ is a smooth map $\gamma : \mathbb{R} \to M$.

Curves can, for example, be thought of as a parametrized version of a path on a manifold with element $t \in [0, 1]$ serving as a time parameter. In this case, if curve goes from point $p \in M$ to point $k \in M$, then $\gamma(0) = p$ and $\gamma(1) = k$.

This allows us to define the tangent vector to the curve $\gamma$ at $p$, which is, essentially, a velocity of curve $\gamma$ at point $p$.

> **Definition 5: Tangent Vector**
>
> Let $\gamma : \mathbb{R} \to M$ be a smooth curve through $p \in M$ and let $\gamma(0) = p$. The **tangent vector** at $p$ along $\gamma$ is the linear map:
>
> $$X_{\gamma,p} : C^\infty(M) \tilde{\to} \mathbb{R}$$
> $$f \mapsto (f \odot \gamma)'(0),$$
>
> where $C^\infty$ is the class of infinitely differentiable functions and $C^\infty(M)$ is a smooth manifold.

Since $f \odot \gamma : \mathbb{R} \to \mathbb{R}$, usual derivation rules can be used.

If we take a closer look at the definition of a tangent vector, it is easy to notice that since there can be infinitely many smooth curves passing through point $p$, there is an infinite amount of tangent vectors defined at $p$. A set of these vectors defines a space, called **tangent space** and is denoted as $T_p M$. In case of a sphere, tangent space can be seen as a hyperplane in Euclidean space, attached to point $p$, which best approximates its neighbourhood, see fig. 2.4b.

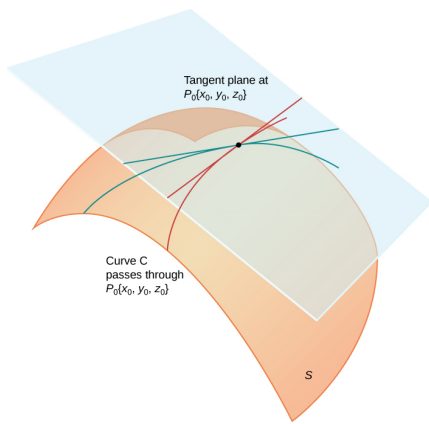

(a) Tangent space and curve of an arbitrary manifold; reproduced from Stewart (2012).

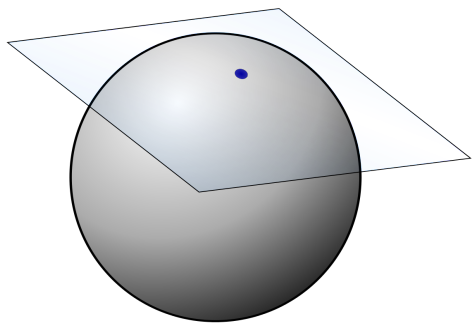

(b) Tangent space of a sphere.

## 2.2 Riemannian Geometry

In the previous sections, we have defined what manifold is and defined vectors over the smooth manifold. These definitions together already provide us with some tooling to operate on these manifolds; however, to allow us to generalize Neural Networks operations to a manifold, we need that it is a Riemannian manifold.

Essentially, Riemannian manifold has an additional structure, called Riemannian metric, that allows us to calculate inner product between two vectors on a manifold, thus allowing for the definition of angles and distances.

> **Definition 6: Riemannian metric (Nakahara, 2003)**
>
> **Riemannian metric** is a family of continuous maps $g_p : T_pM \times T_pM \to \mathbb{R}$ on differentiable manifold $M$ at point $p$, that vary smoothly from point to point, and satisfies following conditions:
> - Symmetry: $g_p(U,V) = g_p(V,U), U, V \in M$
> - Positive-definiteness: $g_p(U,U) \geq 0, U \in M$ with equality holding if and only if $U = 0$

This definition might not look very important at the moment; however, it allows us to define all the essential operations necessary to work with general Riemannian manifolds similarly to how we work with more familiar Euclidean space.

First, we need to start with the definition of affine connection since it will allow us to define geodesic, exponential map, and logarithmic map, which will be discussed in more details further.

> **Definition 7: Affine Connection (Nakahara, 2003)**
>
> **Affine connection** $\nabla$ is a map $\nabla : \mathscr{X}(M) \times \mathscr{X}(M) \to \mathscr{X}(M)$, or $(X,Y) \mapsto \nabla_X Y$, where $X, Y \in \mathscr{X}(M)$, $\mathscr{X}(M)$ is a vector field over smooth manifold $M$, which satisfies following conditions:
> - $\nabla_X(Y + Z) = \nabla_X Y + \nabla_X Z$
> - $\nabla_{(X+Y)} Z = \nabla_X Z + \nabla_Y Z$
> - $\nabla_{(fX)} Y = f \nabla_X Y$
> - $\nabla_X(fY) = X[f]Y + f \nabla_X Y$
>
> where $f$ is a smooth real-valued function defined on manifold $M$ and $X, Y, Z \in \mathscr{X}(M)$

Intuitively, an affine connection can be thought of as a function that, given two vector fields $X$ and $Y$, outputs a third vector field $\nabla_X Y$ that tells us how to transport vector field $Y$ along vector field $X$.

Now, if we define a curve $\gamma(t), t \in [a,b]$ and let $X$ be a vector field defined along the curve and $Y$ - a vector tangent to curve $\gamma$. If $\nabla_Y X = 0 \forall t \in [a,b]$, we call this **parallel transport**. Intuitively, it means that when we move vector field $X$ along curve $\gamma(t)$ all vectors stay the same in terms of length and direction.

Now, if we try using the same setting but parallel transporting tangent field $Y$ along with itself, if $\nabla_Y Y = 0$, it is called **auto-parallel transport**. In this case, curve $\gamma(t)$ is called a **geodesic**. It can be seen a generalization of a straight line in Euclidean space since it is the straightest possible curve in a Riemannian manifold and also represents the shortest path between any two points.

Now, given that we know what a geodesic is, we can define an exponential map and, as its inverse, logarithmic map.

> **Definition 8: Exponential map (Kobayashi and Nomizu, 1996; Fletcher, 2010)**
>
> Given initial conditions $\gamma_v(0) = p$ and $\dot{\gamma}_v(0) = v, v \in T_p M$, **Riemannian exponential map** is defined as:
>
> $$\exp_p : T_p M \to M$$
> $$\exp_p(v) = \gamma_v(1)$$
>
> If affine connection is complete, exponential map is well-defined at every point of the tangent bundle. Otherwise, it is only defined locally.

> **Theorem 2: Exponential map is a local diffeomorphism (Fletcher, 2010)**
>
> Exponential map $\exp_p$ is a diffeomorphism in some neighborhood $V \subset T_p M$ containing 0.

This theorem implies that exponential map has inverse defined, at least locally. It is called **logarithmic map** and is denoted as $\log_p : \exp_p(V) \to V$, where $V \subset T_p M$.

Mainly, exponential and logarithmic maps provide us with a mechanism to map from a vector on a tangent plane to a corresponding point on a manifold and vice versa, as one can see in fig. 2.5.

## 2.3 Hyperbolic Geometry

Finally, it is time to introduce the cornerstone of this work, namely Hyperbolic spaces. First, let us start with an idea of how people arrived at hyperbolic spaces. Euclidean geometry was systematically developed from five main postulates (Walkden, 2012):

1. a straight line may be drawn from any point to any other point
2. a finite straight line may be extended continuously in a straight line

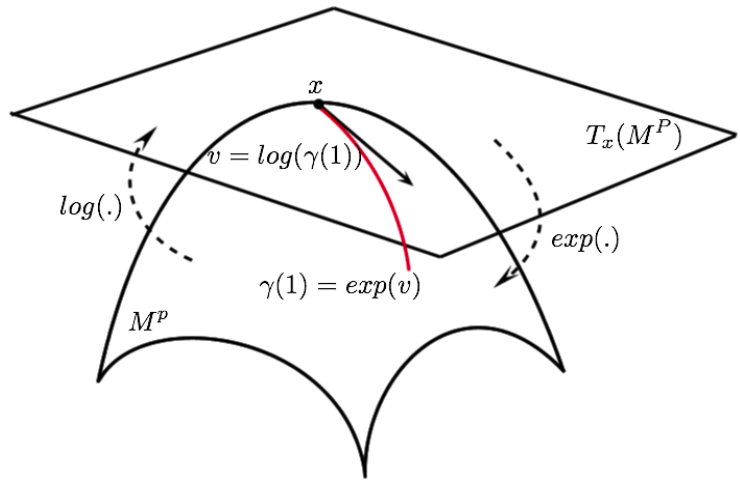

**Fig. 2.5.:** Visualization of an exponential map; reproduced from Simeoni (2013).

3. a circle may be drawn with any centre and any radius

4. all right-angles are equal

5. if a straight line falling on two straight lines makes the interior angles on the same side less than two right angles, then the two straight lines, if extended indefinitely, meet on the side on which the angles are less than two right angles
or, alternatively,
given any infinite straight line and a point not on that line, there exists a unique infinite straight line through that point and parallel to the given line.

While the first four postulates could be accepted without proof, the last one was of significant interest to various scientist throughout many centuries. Most of the people tried to deduce the fifth postulate from the first four. It was left unproven until N. Lobachevsky, who, in 1829, discovered a geometry in which first four postulates would hold but the last one would not. This non-Euclidean geometry is called hyperbolic.

There are multiple different models of hyperbolic spaces; however, since they are identical, we will focus on a Poincaré ball model due to its convenience. In this and the following chapters, we will discuss the main properties that make this model particularly interesting. Note that by saying identical, we mean that there exists an isometry between any two models of hyperbolic geometry.

> **Definition 9: Poincaré ball model**
>
> **Poincaré ball model** $(\mathbb{D}_c^n, g_c^{\mathbb{D}})$ is defined by an underlying manifold $\mathbb{D}_c^n = \{x \in \mathbb{R}^n : c\|x\|^2 < 1\}$ equipped with Riemannian metric:
>
> $$g_c^{\mathbb{D}} = \lambda_x^c g^E,$$
>
> where $\lambda_x^c = \dfrac{2}{1 - c\|x\|^2}$ is the conformal factor
>
> and $g^E = \mathbf{I}_n$ is the Euclidean metric

Note that if $c = 0$, we recover Euclidean space and if $c > 0$, Poincaré ball has a radius of $\frac{1}{\sqrt{c}}$.

> **Note**
>
> In the rest of this work, whenever we say hyperbolic space, we mean a real $d$-dimensional Poincaré ball with $c = 1$ unless explicitly stated otherwise.

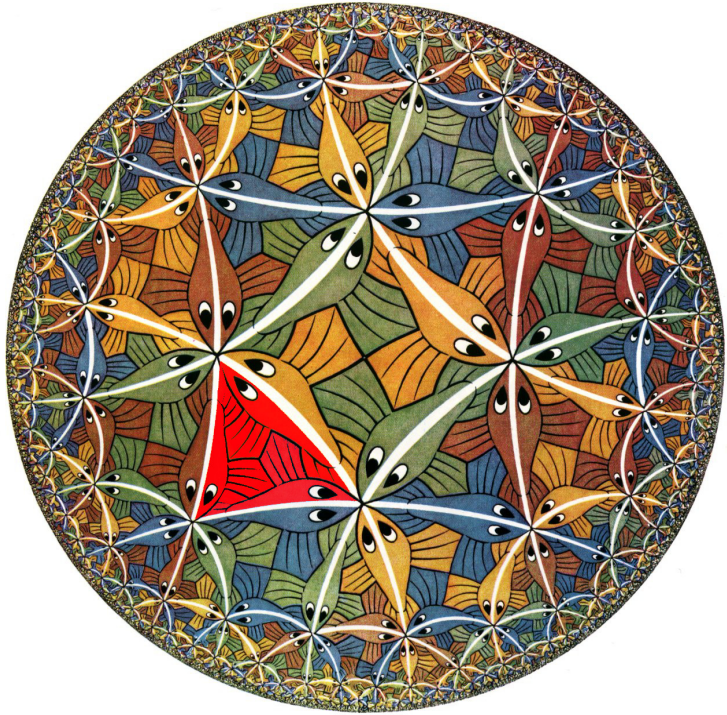

**Fig. 2.6.:** Circle Limit iii, M.C. Escher.
White lines are geodesics and highlighted is a hyperbolic triangle.

Before we proceed with defining main operations on hyperbolic space, it is useful to have a visual understanding of hyperbolic space. Luckily, famous Dutch painter M.C. Escher has provided us with a great visualization of the 2-dimensional Poincaré ball model, see fig. 2.6. An important property that can be seen from this visualization is that hyperbolic spaces can be seen as a continuous version of trees (Hamann, 2011). Gromov's hyperbolicity (Gromov, 1987) was defined to measure how far the graph is from being a tree and, thus, having an underlying hyperbolic geometry. We will discuss Gromov's hyperbolicity in more details in the next sections.

The main reason why hyperbolic spaces can be seen as continuous versions of trees is that the distance from the centre to the point inside the Poincaré ball grows exponentially with the distance between the centre and the border of the ball being infinite. This is consistent with how number of nodes in a tree grows since, a regular tree with branching factor $b$ has $b^l$ nodes at level $l$ and $\frac{(b^l-1)}{b-1}$ nodes on a level less or equal than $l$ (Nickel and Kiela, 2017). Since the area of the surface of $n$-ball in Euclidean space $\mathbb{R}^n$ grows polynomially with radius, the same construction is not possible in Euclidean space. This property of the hyperbolic spaces makes them particularly interesting for embedding hierarchical data which images are one example of.

## 2.4 Gyrovector Spaces

At this point, we could already define all the Riemannian manifold operations for the case of hyperbolic space. However, we will first introduce an elegant framework of gyrovector spaces that will allow us to greatly simplify the derivation of main operations by providing hyperbolic space with space-like vector structure. Before doing so, let us remind ourselves of what a vector space is.

> **Definition 10: Vector Space**
>
> Vector space $\mathbb{V}$ over field $\mathbb{F}$ is a set with two operations:
> - Addition: $+ : \mathbb{V} \times \mathbb{V} \to \mathbb{V}$
> - Scalar multiplication: $\cdot : \mathbb{V} \times \mathbb{F} \to \mathbb{V}$
>
> These operations satisfy following axioms (note that elements from $\mathbb{V}$ are in bold while elements from $\mathbb{F}$ are not):
>
> 1. Associativity of addition:
>    $\mathbf{u} + (\mathbf{v} + \mathbf{w}) = (\mathbf{u} + \mathbf{v}) + \mathbf{w}$
>
> 2. Commutativity of addition:
>    $\mathbf{u} + \mathbf{v} = \mathbf{v} + \mathbf{u}$
>
> 3. Identity element of addition:
>    there exists $\mathbf{0} \in \mathbb{V}$ such that $\mathbf{v} + \mathbf{0} = \mathbf{v} \ \forall \mathbf{v} \in \mathbb{V}$
>
> 4. Inverse element of addition:
>    for every $v \in \mathbb{V}$ there exists an element $-\mathbf{v} \in \mathbb{V}$ such that $\mathbf{v} + (-\mathbf{v}) = \mathbf{0}$
>
> 5. Compatibility of scalar multiplication with field multiplication:
>    $a(b\mathbf{v}) = (ab)\mathbf{v}$
>
> 6. Identity element of scalar multiplication:
>    $1\mathbf{v} = \mathbf{v}$, where 1 denotes multiplicative identity in $\mathbb{F}$.
>
> 7. Distributivity of scalar multiplication with respect to vector addition:
>    $a(\mathbf{u} + \mathbf{v}) = a\mathbf{u} + a\mathbf{v}$
>
> 8. Distributivity of scalar multiplication with respect to field addition:
>    $(a + b)\mathbf{v} = a\mathbf{v} + b\mathbf{v}$

Intuitively, (10) provides us with an abstract definition of vector spaces which can be adapted to arbitrary topological spaces, including familiar Euclidean space. Such generalization, however, might not be straightforward. To have an intuition of why generalization might be problematic, it is useful to take a look at the example of the light cone [1]. As suggested by Minkowski, spacetime can be restricted to hyperbolic geometry and light cone thus also exhibits hyperbolic properties.

Let us start, however, with a more familiar Euclidean space. Imagine moving in a car at a speed of 90 km/h. Next, a car speeds up by 20 km/h. This change of speed can be seen as vector addition. Since we are working in a Euclidean space, vector addition will result in a car having a total speed of 110 km/h.

Now, let us discuss a well-known question of a car moving at speed close to the speed of light. We arbitrarily set it to $0.2c$, where $c$ is the speed of light. Next, passengers decide to turn on headlights, light from which, as postulated by special relativity theory, would move at the speed of light $c$. If we visualize car and light as two vectors and try to add them similarly to how we add Euclidean vectors, we would have the speed of $1.2c$, which higher than such of light. Such a situation is, however, violating special relativity theory and, in reality, the light would still travel at the same speed. This example clearly shows that Euclidean geometry is generally not valid for such spaces.

---

[1] A path that light emanated by a single event would take through spacetime.

To solve the problem described above, Ungar (2005) suggested the theory of gyrovector spaces, which allows us to work with gyrovectors in hyperbolic space similarly to how we would work with vectors in the Euclidean space. As an example, representing car and light as two gyrovectors would not result in an error when adding them together via gyroaddition.

> **Definition 11: Magma**
>
> A set $G$ with a binary operation $\bullet$ defined over it is a **magma** if it satisfies closure axiom:
>
> $$\forall a, b \in G \implies a \bullet b \in G$$

> **Definition 12: Gyrogroup**
>
> A magma $(G, \oplus)$ is a **gyrogroup** if its binary operation satisfies the following axioms:
>
> 1. Left identity element:
>    in $G$ there is at least one element, 0, called a left identity such that $0 \oplus a = a$ for all $a \in G$
>
> 2. Left inverse element:
>    for every $a \in G$ there exists an element $\ominus a \in G$ such that $\ominus a \oplus a = 0$
>
> 3. Left gyroassociativity:
>    for every $a, b, c \in G$ there exists an element $\text{gyr}[a, b](c) \in G$ such that $\oplus$ obeys the left gyroassociative law $a \oplus (b \oplus c) = (a \oplus b) \oplus \text{gyr}[a, b]c$
>
> 4. Gyration $\text{gyr}[a, b] : G \to G$ given by $c \mapsto \text{gyr}[a, b]c$ is an automorphism of $(G, \oplus)$: $\text{gyr}[a, b] \in Aut(G, \oplus)$
>
> 5. Left loop:
>    $\text{gyr}[a, b] = \text{gyr}[a \oplus b, b]$
>
> If also gyrocommutative law $a \oplus b = \text{gyr}[a, b](b \oplus a)$ is obeyed, $(G, \oplus)$ is **gyrocommutative gyrogroup**.

> **Note**
>
> The definition of **gyrogroup** gives rise to the new language, in which we add prefix **gyro-** to show that we are using a hyperbolic version of the definition. As an example, Euclidean space distance has its hyperbolic counterpart, which we will call gyrodistance.

Now we can use this definition for defining gyrovector spaces.

> **Definition 13: Real Inner Product Gyrovector Spaces**
>
> A real inner product gyrovector space $(G, \oplus, \otimes)$ (gyrovector space, in short) is a gyrocommutative gyrogroup $(G, \oplus)$ that obeys the following axioms:
>
> 1. $G$ is a subspace of a real inner product vector space $\mathbb{V}$ called the embedding space of $G$, $G \subset \mathbb{V}$, from which it inherits its inner product, $\cdot$ and norm, $\|\cdot\|$, which are invariant under gyroautomorphisms:
>    $\text{gyr}[\mathbf{u},\mathbf{v}](\mathbf{a}) \cdot \text{gyr}[\mathbf{u},\mathbf{v}](\mathbf{b}) = \mathbf{a} \cdot \mathbf{b}$ for all points $\mathbf{a}, \mathbf{b}, \mathbf{u}, \mathbf{v} \in G$
>
> 2. $G$ admits a scalar multiplication, $\otimes$, possessing the following properties. For all scalars (real numbers) $r, r_1, r_2 \in \mathbb{R}$ and all points $\mathbf{a} \in G$:
>    a) Identity Scalar Multiplication: $1 \otimes \mathbf{a} = \mathbf{a}$
>    b) Scalar Distributive Law: $(r_1 + r_2) \otimes \mathbf{a} = r_1 \otimes \mathbf{a} \oplus r_2 \otimes \mathbf{a}$
>    c) Scalar Associative Law: $(r_1 r_2) \otimes \mathbf{a} = r_1 \otimes (r_2 \otimes \mathbf{a})$
>    d) Scaling Property: $\frac{|r| \otimes \mathbf{a}}{\|r \otimes \mathbf{a}\|} = \frac{\mathbf{a}}{\|\mathbf{a}\|}, \mathbf{a} \neq \mathbf{0}, r \neq 0$
>    e) Gyroautomorphism Property: $\text{gyr}[\mathbf{u},\mathbf{v}](r \otimes \mathbf{a}) = r \otimes \text{gyr}[\mathbf{u},\mathbf{v}](\mathbf{a})$
>    f) Identity Gyroautomorphism: $\text{gyr}[r_1 \otimes \mathbf{v}, r_2 \otimes \mathbf{v}] = I$
>
> 3. Real, one-dimensional vector space structure $(\|G\|, \oplus, \otimes)$ for the set $\|G\|$ of one-dimensional vectors:
>    $\|G\| = \pm \|\mathbf{a}\| : \mathbf{a} \in G \subset \mathbb{R}$
>    with vector addition $\oplus$ and scalar multiplication $\otimes$, such that for all $r \in \mathbb{R}$ and $\mathbf{a}, \mathbf{b} \in G$:
>    a) Homogeneity Property: $\|r \otimes \mathbf{a}\| = |r| \otimes \|\mathbf{a}\|$
>    b) Gyrotriangle Inequality: $\|\mathbf{a} \oplus \mathbf{b}\| \leq \|\mathbf{a}\| \oplus \|\mathbf{b}\|$

From the gyrovector space axioms, next theorem follows.

> **Theorem 3: Gyrovector space properties**
>
> Let $(G, \oplus, \otimes)$ be a gyrovector space whose embedding space is a real inner product vector space $\mathbb{V}$, and let $0$, $\mathbf{0}$ and $\mathbf{0}_\mathbb{V}$ be the neutral elements of the real line $(\mathbb{R}, +)$, the gyrocommutative gyrogroup $(G, \oplus)$, and the vector space $(\mathbb{V}, +)$ respectively. Then, for all $n \in \mathbb{N}$, $r \in \mathbb{R}$ and $\mathbf{a} \in G$:
>
> 1. $0 \otimes \mathbf{a} = \mathbf{0}$
> 2. $n \otimes \mathbf{a} = \underbrace{\mathbf{a} \oplus \ldots \oplus \mathbf{a}}_{n}$
> 3. $(-r) \otimes \mathbf{a} = \ominus(r \otimes \mathbf{a})$
> 4. $r \otimes \mathbf{0} = \mathbf{0}$
> 5. $r \otimes (\ominus \mathbf{a}) = \ominus(r \otimes \mathbf{a})$
> 6. $\|\ominus \mathbf{a}\| = \|\mathbf{a}\|$
> 7. $\mathbf{0} = \mathbf{0}_\mathbb{V}$
> 8. $r \otimes \mathbf{a} = \mathbf{0} \iff (r = 0 \text{ or } \mathbf{a} = \mathbf{0})$

Before we can proceed to a concrete example of a gyrovector space, we would like to define necessary additional operations first.

> **Definition 14: Gyrodistance**
>
> Let $(G, \oplus, \otimes)$ be a gyrovector space whose ambient space is a real inner product vector space $\mathbb{V}$. Its gyrometric is given by the **gyrodistance** function $d_\oplus(\mathbf{a}, \mathbf{b}) : G \times G \to \mathbb{R}^{\geq 0}$:
>
> $$d_\oplus(\mathbf{a}, \mathbf{b}) = \|\ominus \mathbf{a} \oplus \mathbf{b}\| = \|\mathbf{b} \ominus \mathbf{a}\|,$$
>
> where $d_\oplus(\mathbf{a}, \mathbf{b})$ is the gyrodistance of $\mathbf{a}$ to $\mathbf{b}$.

Gyrodistance, as we have already discussed in part about an affine connection ((7)) is the length of the geodesic connecting two points. Geodesics in the context of gyrovector spaces are called **gyrolines**. Similarly to the general definition of geodesic, we rely on the notion of time to define gyroline.

> **Definition 15: Gyroline**
>
> Let $\mathbf{a}, \mathbf{b}$ be any two distinct points in a gyrovector space $(G, \oplus, \otimes)$. The gyroline in $G$ that passes through the points $\mathbf{a}$ and $\mathbf{b}$ is the set of all points:
>
> $$L_{\mathbf{a} \to \mathbf{b}}(t) = \mathbf{a} \oplus (\ominus \mathbf{a} \oplus \mathbf{b}) \otimes t$$
>
> in $G$ with $t \in \mathbb{R}$.

Note that this definition is consistent with its more familiar Euclidean counterpart. At $t = 0 : \mathbf{a} \oplus (\ominus \mathbf{a} \oplus \mathbf{b}) \otimes t = \mathbf{a}$, while at $t = 1 : \mathbf{a} \oplus (\ominus \mathbf{a} \oplus \mathbf{b}) \otimes t = \mathbf{b}$. When $t = \frac{1}{2}$, we naturally arrive at the definition of gyromidpoint.

> **Definition 16: Gyromidpoint**
>
> The gyromidpoint $\mathbf{m}_{\mathbf{ac}}$ of any two distrinct points $\mathbf{a}$ and $\mathbf{c}$ in a gyrovector space $(G, \oplus, \otimes)$ is given by the equation:
>
> $$\mathbf{m}_{\mathbf{ac}} = \mathbf{a} \oplus (\ominus \mathbf{a} \oplus \mathbf{c}) \otimes \frac{1}{2}$$

Gyromidpoints exhibits properties similar to those of a more familiar Euclidean midpoint. Namely, symmetry is preserved $\mathbf{m}_{\mathbf{ac}} = \mathbf{m}_{\mathbf{ca}}$.

> **Note**
>
> A fundamental property of gyromidpoint is that it can be generalized to a case of more than two points $(\mathbf{a}_1, \mathbf{a}_2, \ldots \mathbf{a}_n)$. In this case, $\mathbf{m}_{\mathbf{a}_1 \mathbf{a}_2 \ldots \mathbf{a}_n}$ can be seen as a hyperbolic version of a centroid, or, in statistical terms, arithmetic mean.

Now, we are ready to define Möbius gyrovector space that forms the algebraic setting for the Poincaré ball model of hyperbolic geometry.

> **Definition 17: Möbius gyrovector space**
>
> **Möbius gyrovector space** is a gyrovector space $(\mathbb{D}_c^n, \oplus_c, \otimes_c)$ with $\oplus_c : \mathbb{D}_c^n \times \mathbb{D}_c^n \to \mathbb{D}_c^n$ and $\otimes_c : \mathbb{D}_c^n \otimes \mathbb{R} \to \mathbb{D}_c^n$ defined as:
>
> $$\mathbf{u} \oplus_c \mathbf{v} = \frac{(1 + 2c\mathbf{u} \cdot \mathbf{v} + c\|\mathbf{v}\|^2)\mathbf{u} + (1 - \|\mathbf{u}\|^2)\mathbf{v}}{1 + 2c\mathbf{u} \cdot \mathbf{v} + c^2\|\mathbf{u}\|^2\|\mathbf{v}\|^2}$$
>
> $$r \otimes_c \mathbf{u} = \frac{1}{\sqrt{c}} \tanh\left(r \tanh^{-1}(\sqrt{c}\|\mathbf{u}\|)\right) \frac{\mathbf{u}}{\|\mathbf{u}\|}$$

One can prove that both of the operations satisfy gyrovector space axioms. As expected, $\otimes_c$ has priority over $\oplus_c$.

Apart from providing us with an essential way to define further necessary operations, (17) is also consistent with previously defined gyrooperations, see fig. 2.7a for visual understanding of Möbius gyroline.

> **Definition 18: Möbius distance**
>
> Möbius gyrovector space $(\mathbb{D}_c^n, \oplus_c, \otimes_c)$ is equipped with distance metric $d_c(\mathbf{u}, \mathbf{v}) : \mathbb{D}_c^n \times \mathbb{D}_c^n \to \mathbb{R}^{\geq 0}$:
>
> $$d_c(\mathbf{u}, \mathbf{v}) = \frac{2}{\sqrt{c}} \tanh^{-1}(\sqrt{c} \| -\mathbf{u} \oplus_c \mathbf{v} \|)$$

> **Definition 19: Möbius midpoint**
>
> Möbius gyrovector space $(\mathbb{D}_c^n, \oplus_c, \otimes_c)$ is equipped with a midpoint operation $f : (\mathbf{a}_1, \mathbf{a}_2, \ldots \mathbf{a}_n) \mapsto \mathbf{m}_{\mathbf{a}_1 \mathbf{a}_2 \ldots \mathbf{a}_n}$, where $\mathbf{a}_{1\ldots n} \in \mathbb{D}_c^n$:
>
> $$\mathbf{m}_{\mathbf{a}_1 \mathbf{a}_2 \ldots \mathbf{a}_n} = \frac{1}{2} \otimes_c \varphi^{-1} \left( \frac{\sum_{i=1}^n 2\gamma_{\mathbf{a}_i}^2 \varphi(\mathbf{a}_i)}{\sum_{i=1}^n (2\gamma_{\mathbf{a}_i}^2 - 1)} \right),$$
>
> where $\gamma_{\mathbf{a}_i} = \dfrac{1}{\sqrt{1 - c\|\mathbf{a}_i\|^2}}$ is a Lorentz factor,
>
> $\varphi : \mathbb{D}_c^n \to \mathbb{R}^n$ is a mapping from the hyperbolic space to its embedding space

An essential extension to the Möbius midpoint is weighted Möbius midpoint, which corresponds to weighted mean in Euclidean space.

> **Definition 20: Möbius weighted midpoint**
>
> Möbius gyrovector space $(\mathbb{D}_c^n, \oplus_c, \otimes_c)$ is equipped with a weighted midpoint operation $f : (\mathbf{a}_1, \mathbf{a}_2, \ldots \mathbf{a}_n, \alpha_1, \alpha_2, \ldots \alpha_n) \mapsto \mathbf{m}_{\mathbf{a}_1 \mathbf{a}_2 \ldots \mathbf{a}_n, \alpha_1, \alpha_2, \ldots \alpha_n}$, where $\mathbf{a}_{1\ldots n} \in \mathbb{D}_c^n, \alpha_{1\ldots n} \in \mathbb{R}$:
>
> $$\mathbf{m}_{\mathbf{a}_1 \mathbf{a}_2 \ldots \mathbf{a}_n, \alpha_1, \alpha_2, \ldots \alpha_n} = \frac{1}{2} \otimes_c \varphi^{-1} \left( \frac{\sum_{i=1}^n 2\alpha_i \gamma_{\mathbf{a}_i}^2 \varphi(\mathbf{a}_i)}{\sum_{i=1}^n (2\alpha_i \gamma_{\mathbf{a}_i}^2 - 1)} \right),$$
>
> where $\gamma_{\mathbf{a}_i} = \dfrac{1}{\sqrt{1 - c\|\mathbf{a}_i\|^2}}$ is a Lorentz factor,
>
> $\varphi : \mathbb{D}_c^n \to \mathbb{R}^n$ is a mapping from the hyperbolic space to its embedding space

> **Note 1**
>
> In the rest of this work, whenever Möbius midpoint or its weighted version is used, mapping function $\varphi : \mathbb{D}_c^n \to \mathbb{R}^n$ is implicit and, therefore, omitted. Therefore, in the rest of this work:
>
> $$\mathbf{m}_{\mathbf{a}_1\mathbf{a}_2\dots\mathbf{a}_n} = \frac{1}{2} \otimes_c \varphi^{-1} \left( \frac{\sum_{i=1}^n 2\gamma_{\mathbf{a}_i}^2 \varphi(\mathbf{a}_i)}{\sum_{i=1}^n (2\gamma_{\mathbf{a}_i}^2 - 1)} \right)$$
>
> is equivalent to
>
> $$\mathbf{m}_{\mathbf{a}_1\mathbf{a}_2\dots\mathbf{a}_n} = \frac{1}{2} \otimes_c \frac{\sum_{i=1}^n 2\gamma_{\mathbf{a}_i}^2 \mathbf{a}_i}{\sum_{i=1}^n (2\gamma_{\mathbf{a}_i}^2 - 1)}$$

Finally, since Poincaré ball is a Riemannian manifold, we can connect Riemannian geometry and Hyperbolic geometry using gyrovector spaces, see fig. 2.7.

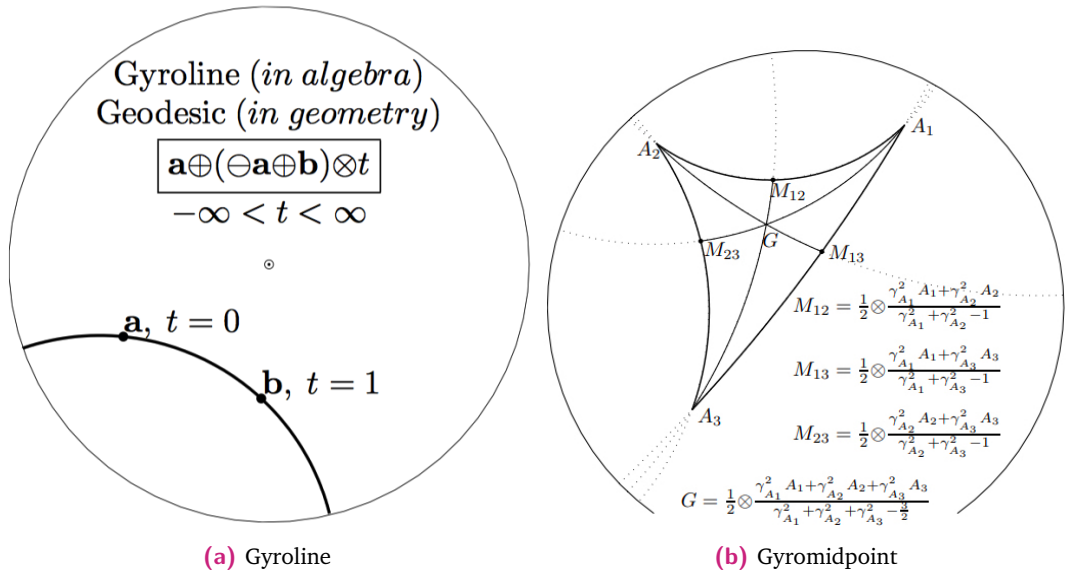

(a) Gyroline

(b) Gyromidpoint

**Fig. 2.7.:** Visualization of a gyroline and a gyromidpoint on a Poincaré disk; reproduced from Ungar (2005).

> **Definition 21: Riemannian operations in Möbius gyrovector space**
>
> Möbius gyrovector space $(\mathbb{D}_c^n, \oplus_c, \otimes_c)$ is a Riemannian manifold and, therefore, has following operations defined over it:
>
> $$\text{Exponential map: } \exp_\mathbf{u}^c(\mathbf{v}) = \mathbf{u} \oplus_c \left( \tanh\left( \sqrt{c} \frac{\lambda_\mathbf{u}^c \|\mathbf{v}\|}{2} \right) \frac{\mathbf{v}}{\sqrt{c}\|\mathbf{v}\|} \right), \ \mathbf{v} \in T_\mathbf{u} \mathbb{D}_c^n$$
>
> $$\text{Zero exponential map: } \exp_\mathbf{0}^c(\mathbf{v}) = \tanh(\sqrt{c}\|\mathbf{v}\|) \frac{\mathbf{v}}{\sqrt{c}\|\mathbf{v}\|}, \ \mathbf{v} \in T_\mathbf{0} \mathbb{D}_c^n$$
>
> $$\text{Logarithmic map: } \log_\mathbf{u}^c(\mathbf{v}) = \frac{2}{\sqrt{c}\lambda_\mathbf{u}^c} \tanh^{-1}(\sqrt{c}\|-\mathbf{u} \oplus_c \mathbf{v}\|) \frac{-\mathbf{u} \oplus_c \mathbf{v}}{\|-\mathbf{u} \oplus_c \mathbf{v}\|}, \ \mathbf{v} \in \mathbb{D}_c^n$$
>
> $$\text{Zero logarithmic map: } \log_\mathbf{0}^c(\mathbf{v}) = \tanh^{-1}(\sqrt{c}\|\mathbf{v}\|) \frac{\mathbf{v}}{\sqrt{c}\|\mathbf{v}\|}, \ \mathbf{v} \in \mathbb{D}_c^n$$
>
> $$\text{Parallel transport: } P_{\mathbf{0} \to \mathbf{u}}^c(\mathbf{v}) = \log_\mathbf{u}^c(\mathbf{u} \oplus_c \exp_\mathbf{0}^c(\mathbf{v})), \ \mathbf{v} \in T_\mathbf{u} \mathbb{D}_c^n$$

Interestingly, exponential map and logarithmic map provide us with an alternative view of the Möbius gyrovector space operations, which we will heavily use when redefining Neural Networks operations to their hyperbolic case.

> **Theorem 4: Möbius operations using exponential and logarithmic maps (Ganea et al., 2018a)**
>
> Möbius gyrovector space $(\mathbb{D}_c^n, \oplus_c, \otimes_c)$ operations can be redefined using exponential and logarithmic maps as following:
>
> $$\text{Möbius scalar multiplication: } r \otimes_c \mathbf{u} = \exp_\mathbf{0}^c(r \log_\mathbf{0}^c(\mathbf{u}))$$
>
> $$\text{Gyroline: } L_{\mathbf{a} \to \mathbf{b}}(r) = \exp_\mathbf{a}^c(r \log_\mathbf{a}^c(\mathbf{b}))$$
>
> $$\text{Parallel Transport: } P_{\mathbf{0} \to \mathbf{u}}^c(\mathbf{v}) = \frac{\lambda_\mathbf{0}^c}{\lambda_\mathbf{u}^c} \mathbf{v}$$

There are two strikingly important moments about this theorem. First of all, similarly to the (9), if $c \to 0$, all Möbius operations turn into their Euclidean counterparts, which provides us with a mechanism to let model choose which space is preferable for the task. Secondly, it allows us to see that using hyperbolic versions of Euclidean operations amounts to mapping to the Euclidean space, performing an operation there and then mapping back to the Hyperbolic space. This observation will be more apparent when we discuss work by Ganea et al. (2018a) in the next chapter.

## 2.5 Gromov Hyperbolicity

As a final part of this chapter, we would like to introduce the notion of Gromov Hyperbolicity (Gromov, 1987), which allows us to classify graphs based on how well they can be embedded in the hyperbolic space. Gromov hyperbolicity will come extremely useful when discussing the performance of different graph embedding models in the following chapters.

In his work on Hyperbolic Group Theory, Gromov has introduced an important concept of hyperbolic metric space and hyperbolic groups. Introduced in 1987, Gromov's hyperbolic metric space has quickly become a popular tool to analyze the structure of graphs with applications in routing, navigation, and visualization of the Internet. (Cvetkovski and Crovella, 2009; Boguna et al., 2009; Jonckheere and Lohsoonthorn, 2004; Kleinberg, 2006).

Using Gromov's hyperbolic metric space is being complicated by the fact that it has several equivalent definitions (up to a multiplicative constant). However, in this work, we will use the following definition.

> **Definition 22: Gromov's four point condition (Gromov, 1987))**
>
> In a metric space $(X, d)$, given $u, v, w, x$ with $d(u,v) + d(w,x) \geq d(u,x) + d(w,v) \geq d(u,w) + d(v,x)$ in $X$, we note hyperbolicity $\delta(u, v, w, x) = \frac{(d(u,v)+d(w,x)-d(u,x)-d(w,v))}{2}$.
> $(X, d)$ is called $\delta$-hyperbolic for some non-negative real number $\delta$ if for any four points $u, v, w, x \in X, \delta(u, v, w, x) \leq \delta$. Let $\delta(X, d)$ be the smallest possible value of such $\delta$, which can also be defined as $\delta(X, d) = sup_{u,v,w,x \in X} \delta(u, v, w, x)$.

An undirected, unweighted and connected graph $G = (V, E)$ can be interpreted as a hyperbolic metric space $(V, d_G)$, where $d_G(u, v)$ is the graph distance[2] between two vertices $u$ and $v$. This allows us to apply (22) to it, thus getting hyperbolicity $\delta(G, d_G)$ as a global property of the graph $G$. As we will soon see, this characterizes how close the graph is to be a tree and, thus, how well it can be embedded in the hyperbolic space.

This is motivated by the following: $\delta$ for an arbitrary graph is bound between $\frac{D(G)}{2}$, where $D(G)$ is the diameter[3] of the graph, and 0. While trees and block graphs are 0-hyperbolic, and, as discussed before, can be entirely embedded in the hyperbolic space with low distortion, $n \times n$ grids are $\frac{n-1}{2}$-hyperbolic and, thus, cannot be embedded into hyperbolic space with arbitrarily low distortion.

---
[2]The neighboring nodes have a distance of 1.
[3]Distance between two furthest vertices in a graph.

Although we have mentioned certain graphs along with their hyperbolicity, before we proceed, it would be helpful to revise main types of graphs to see how they can be interpreted from the hyperbolic metric spaces perspective. As suggested in Frogner et al. (2019), we will use scale-free networks, small-world networks, community-structured networks, and trees. We will generate each one of them randomly measuring hyperbolicity for a different number of vertices as well as connectivity. Examples of each type of graph can be seen in fig. 2.8

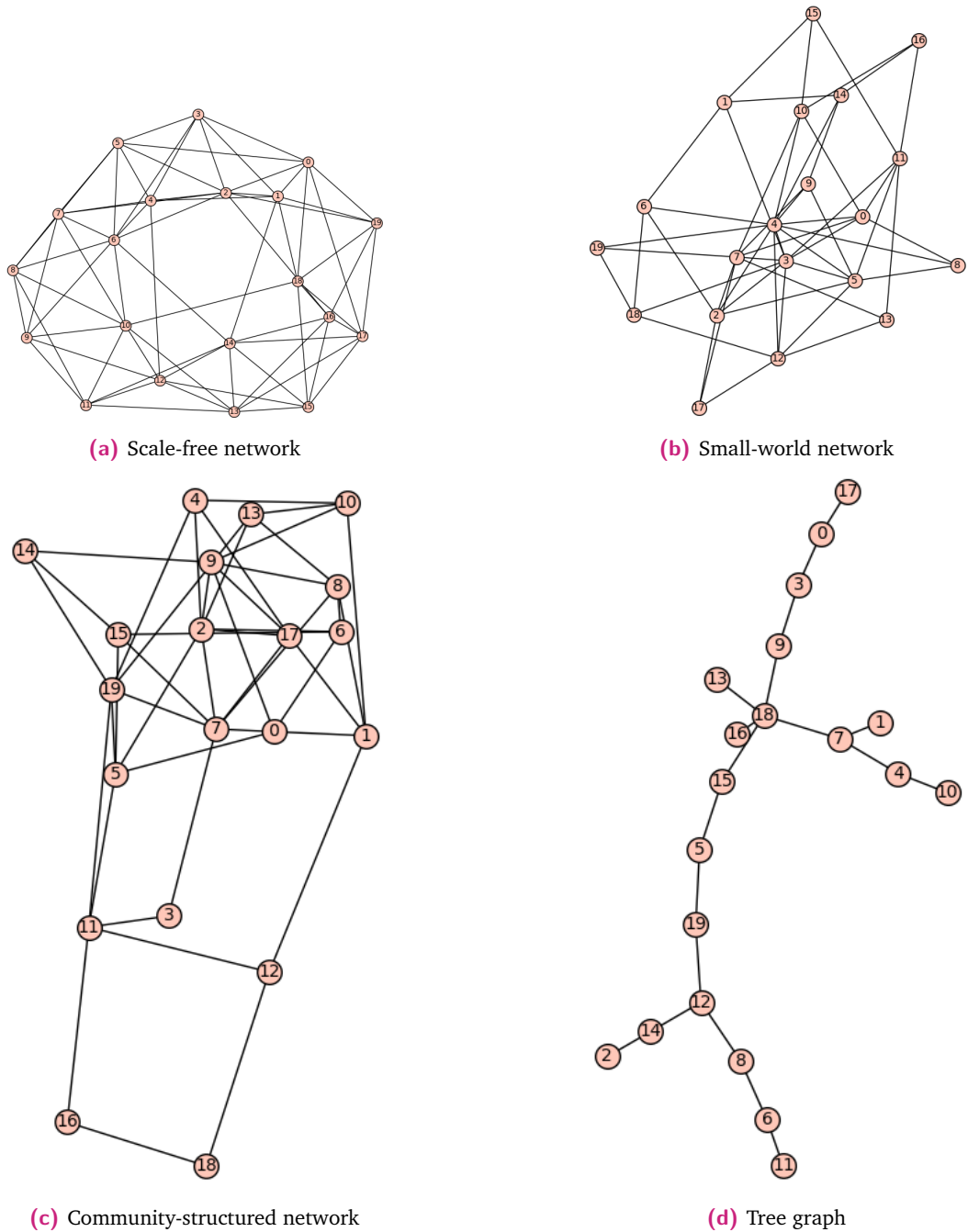

**Fig. 2.8.:** Visualization of different types of graphs.

Scale-free networks are characterized by having degree distribution, which follows a power law. What this means is that the number $P(k)$ of vertices in the network having $k$ connections to other vertices goes for large values of $k$ as $P(k) \sim k^{-\gamma}$, $\gamma$ typically being in the range of $[2, 3]$. This leads to a property that distances between vertices are scaling logarithmically with the number of vertices. Unsurprisingly, a lot of real-world networks, e.g., citation networks, social networks, are scale-free. To generate these randomly, we use the Barabasi-Albert model (Barabási and Albert, 1999), which takes as parameters number of vertices and number of edges to attach to each vertex. We generate graphs with $V = 2^k, k \in [7, 15]$ being the number of vertices and $M = 5 \cdot k, k \in [1, 5]$ being the attachment parameter. We use 5 being the h-index of Maurice Weiler[4] as a starting point since h-index is famously scale-free.

Small-world networks are networks in which nodes are not neighbours of each other but are likely to have neighbours in common. This leads to that every node can be reached from any other node by a small number of hops. More formally, small-world networks have a low average path length, which is the main characteristic of the small-world networks. There have been experiments (Dodds et al., 2003) demonstrating that the world indeed possesses this property and outside of the mathematical community, it is widely known as six degrees of separation (Wikipedia contributors, 2019). Small-World networks can be seen as scale-free networks with the additional property of having distinct connected clusters. To generate these randomly, we use Newmann-Watts-Strogatz model (Newman et al., 2002), which takes as parameters number of vertices and number of neighbours for each vertex as well as probability for adding a random edge between each pair of non-connected vertices. We use parameters as suggested in the original paper and generate graphs with $V = 2^k, k \in [7, 15]$ being the number of vertices and $M = \ln(V) \cdot k, k \in [2, 5]$ being the number of neighbours, $p = 0.15$ being the probability of adding random edge.

Community-structured networks are generated from the stochastic block model, which first generates densely connected communities and then randomly generates sparse connections between each possible pair of them. It takes as parameters number of vertices and number of communities to split those vertices into. Again, we use $V = 2^k, k \in [7, 15]$ being the number of vertices and ensure that each community has between $\ln(V)$ and 15 members. This results in $M \in [\frac{V}{15}, \frac{V}{\ln(V)}]$, which we split into 5 parts.

Finally, we generate random trees with $V = 2^k, k \in [7, 15]$ vertices.

Next, we use SageMath (The Sage Developers, 2019) to measure hyperbolicity in the generated graphs. Since calculating hyperbolicity is $O(n^4)$, where $n$ is the number of vertices, we use an improvement over the naive algorithm, suggested in Borassi et al. (2015). The suggested algorithm uses the notion of nodes being 'far-apart' and additional heuristics to remove some of the four-tuples from the comparison. This significantly speeds up the algorithm, especially for large graphs, albeit at the expense of higher memory cost. The results of our measurements are visualized in fig. 2.9.

---

[4] https://scholar.google.com/citations?user=uQePx6EAAAAJ&hl=en

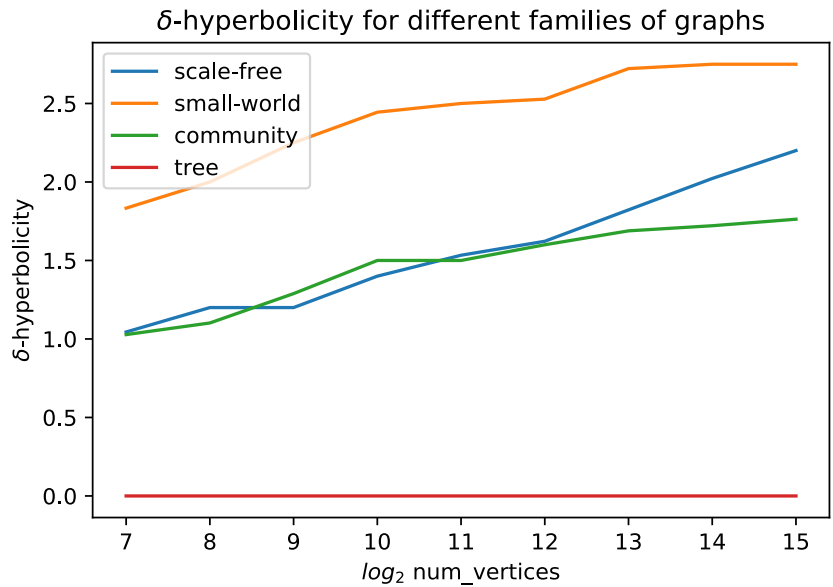

**Fig. 2.9.:** Hyperbolicity for different types of graphs.

We can see that with the number of nodes, all networks except for trees become less and less hyperbolic, while trees preserve the same constant hyperbolicity. Aforementioned is in line with our understanding of hyperbolicity since all networks we have observed get sparser as they grow, which leads to higher hyperbolicity, while the same cannot be said about trees. Interestingly enough, hyperbolicity of community-structured graphs grows the slowest thus making them the most hyperbolic among all non-tree graphs (at least asymptotically), which can be explained by the fact that they are the closest to the block graphs being constructed by sparsifying those.

This result is of high importance for this work for two reasons:

- It provides us with a theoretical framework for explaining the performance of hyperbolic graph embedding models. We will use it in the next chapter when discussing the results of Frogner et al. (2019);
- It raises a lot of questions worth future research, which we will talk about in-depth in section 7.1.

Current subsection concludes an overview of the main mathematical ideas that we will use throughout this work. We will now proceed to a discussion of related works.

# 3 Literature Review

In this chapter, we provide an overview of recent works using hyperbolic spaces that are of high relevance for this thesis.

## 3.1 Hyperbolic Graph Embeddins

Now, we would like to discuss a topic of Hyperbolic Graph Embeddings, which is of crucial importance to this work. We will mostly focus on works by Nickel and Kiela (2017) and Ganea et al. (2018b), however, a brief discussion of recent work by Frogner et al. (2019) will be held.

### 3.1.1 WordNet hierarchy embedding

Paper by Nickel and Kiela (2017) came to be the first paper to touch upon the topic of using Hyperbolic space for machine learning tasks, namely, network embedding in this case. Essentially, the paper suggests that using manifold that reflects the structure of the graph that we are embedding can lead to lower embedding distortion and, thus, better performance of machine learning models. This suggestion is consistent with Whitney's theorem, specified in (1) and experimental results indeed support this statement.

Paper is mostly concerned with optimally embedding hierarchies. Since, as previously discussed in section 2.3, hyperbolic space is the most suitable with trees, it also provides a very convenient space for embedding hierarchies[5]. Therefore, by using Poincaré ball, authors of the paper intend to solve two issues with previously suggested methods for hierarchy embeddings (Vendrov et al., 2015; Vilnis and McCallum, 2014):

1. Learn better embeddings by inducing a suitable bias on the form of the embedding space;
2. Get additional insights into relationships between different elements of the hierarchy by capturing it explicitly in the embedding space.

---
[5]Since hierarchies have tree-like structure

Indeed, using hyperbolic distance, as defined in (18) allows us to solve both of these problems since it can simultaneously (a) capture hierarchical relationships of the objects through distance from the origin[6] and (b) capture similarity between nodes in a hierarchy through the distance between nodes of interest.

Authors suggest including hyperbolic distance through a task-specific loss function with Euclidean distance replaced by a Moebius distance. They discuss two tasks, which can be performed on the graphs, namely, graph reconstruction and link prediction. Graph reconstruction task is used to evaluate representation capacity while link prediction is used to test generalization performance.

Soft ranking loss is used for the task of graph reconstruction and is defined as following:

$$L(\theta) = \sum_{(u,v) \in \mathbb{G}} \log \frac{e^{-d(u,v)}}{\sum_{v' \in \mathbb{N}(u)} e^{-d(u,v')}},$$

where $mathbbG = \{(u,v)\}$ is the set of observed relations, $\mathbb{N}(u)$ is the set of negative examples

For the case link prediction, authors use Fermi-Dirac distribution for edge probability

$$P((u,v) = 1|\theta) = \frac{1}{e^{\frac{(d(u,v)-r)}{t}} + 1}$$

where $r, t > 0$ are hyperparameters. In this distribution, $r$ is the radius around each point $u$ within which neighboring nodes are likely to lie and $t$ is the steepness of the logistic function. This distribution, combined with cross-entropy yields the final loss.

Poincaré embeddings have been shown to perform well on both of these tasks, outperforming all the existing algorithm at a time by a large margin, as one can see in table 3.1. An important observation, however, is that, as dimensionality increases, the margin between Euclidean and Poincaré embeddings gets smaller. Reasons for this will be the main point of discussion in the section 3.1.2.

Although showing excellent performance, suggested an approach for graph reconstruction suffered from multiple flaws, which were discussed and further solved in work by Ganea et al. (2018b). First of all, the suggested approach was not capable of encoding asymmetric relations due to the symmetry of distance function. Secondly, as observed experimentally, it led to points collapsing on the border of the Poincaré ball. By suggest-ing a better loss function, authors were able to solve both of these issues, which led to much better performance, as can be seen in fig. 3.1 and table 3.2.

---

[6] With origin node lying at the origin of embedding space $\mathbb{R}^n$.

|               |              | Dimensionality |       |       |       |       |       |
|---------------|--------------|-------|-------|-------|-------|-------|-------|
|               |              | 5     | 10    | 20    | 50    | 100   | 200   |
| Reconstruction | Euclidean    | 0.024 | 0.059 | 0.087 | 0.14  | 0.162 | 0.168 |
|               | Translational | 0.517 | 0.503 | 0.563 | 0.566 | 0.562 | 0.565 |
|               | Poincaré     | 0.823 | 0.851 | 0.855 | 0.86  | 0.857 | 0.87  |
| Link Prediction | Euclidean  | 0.024 | 0.059 | 0.176 | 0.286 | 0.428 | 0.49  |
|               | Translational | 0.545 | 0.554 | 0.554 | 0.56  | 0.562 | 0.559 |
|               | Poincaré     | 0.825 | 0.852 | 0.861 | 0.863 | 0.856 | 0.855 |

**Tab. 3.1.:** Mean Average Precision on tasks of Reconstruction and Link Prediction. Euclidean embeddings only use Euclidean distance as a loss, while Translational stands for TransE model suggested in (Bordes et al., 2013)

| Method           | Dimension = 5 | Dimension = 10 |
|------------------|---------------|----------------|
| Poincaré         | 83.6%         | 85.3%          |
| Hyperbolic Cones | 92.8%         | 94.4%          |

**Tab. 3.2.:** F1 score on the graph reconstruction task (Ganea et al., 2018b). Poincaré is the method suggested by Nickel and Kiela (2017), Hyperbolic Cones is suggested by Ganea et al. (2018b).

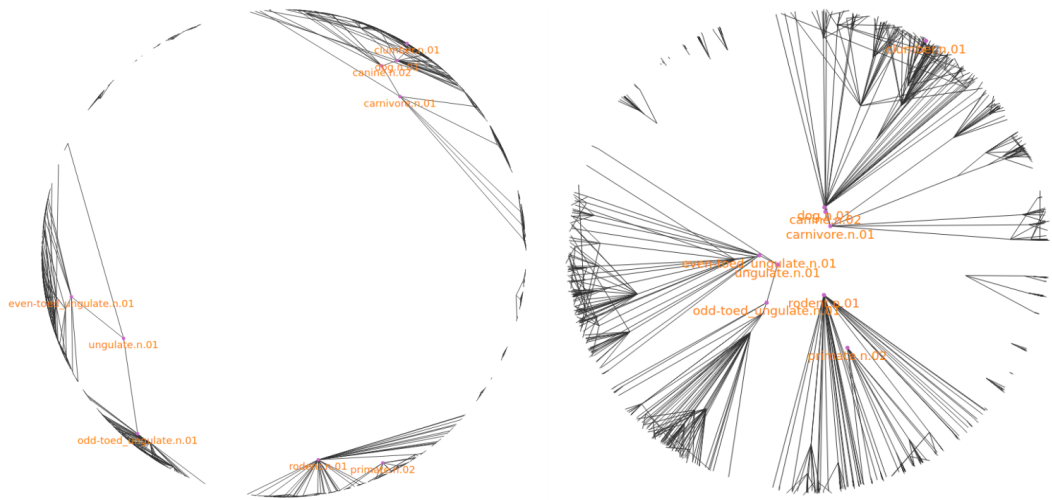

**Fig. 3.1.:** Visualization of 2D embedding of WordNet mammal subtree using (Nickel and Kiela, 2017) method (left) and (Ganea et al., 2018b) method (right). Qualitative evaluation clearly points at that right image is better at embedding WordNet hierarchy; reproduced from Ganea et al. (2018b).

### 3.1.2 General Graphs Embedding

The question that naturally arises while studying hyperbolic spaces as embedding space for graphs is whether they are suitable for embedding arbitrary graphs, not just trees. Previous results have demonstrated that most of the graphs can be embedded better when assuming an underlying hyperbolic geometry (Cvetkovski and Crovella, 2009; Cvetkovski and Crovella, 2009; Zhao et al., 2011; Krioukov et al., 2010). However, there have been several papers (Chen et al., 2012; Frogner et al., 2019) theoretically and empirically, proving that it is only the case for trees and Euclidean geometry is more suitable for any other type of graphs. These two statements contradict each other, while the first statement also contradicts Gromov Hyperbolicity discussed in section 2.5. Therefore, since one of the goals of this work is studying whether GCNs can benefit from using hyperbolic space, understanding which graphs benefit from hyperbolic space is of paramount importance.

The paper of particular interest in our work is the one by Frogner et al. (2019). Unlike Chen et al. (2012), who provides a theoretical justification of why small world graphs can't be embedded with low distortion in the hyperbolic space, it (a) empirically shows that hyperbolic embedding is only advantageous for trees and (b) provides an empirical reasoning of why previous papers have found graphs to have approximately hyperbolic underlying geometry. Although the task of the paper is different, namely, showing that Wasserstein spaces are suitable for embedding most of the types of graphs, we would like to focus on its evaluation of graph embedding for different dimensionalities of the hyperbolic space, see fig. 3.2.

From fig. 3.2 we can see that hyperbolic space only outperforms Euclidean space on random trees or, for other network structure, in the low dimensional setting. This result is compatible with our understanding since we know that hyperbolic space can, indeed, embed trees with arbitrarily low distortion. However, the reason for performing well in the low-dimensional setting is due to the following observation: the capacity of space necessary to embed the hierarchy is limited, i.e., we can embed a finite hierarchy in the finite-dimensional space with arbitrarily low distortion, assuming that we can use space of very high dimensionality. Therefore, since the capacity of a Poincaré ball is much higher than such of Euclidean space in the lower-dimensional setting, we can observe a considerable margin between the performance of the model. However, as dimensionality increases, the capacity of the Euclidean space gets sufficient for embedding the WordNet hierarchy, thus getting closer in terms of performance to the Poincaré model.

Importantly, however, when comparing their results with the ones from (Cvetkovski and Crovella, 2009; Cvetkovski and Crovella, 2009; Zhao et al., 2011; Krioukov et al., 2010), we can notice that all the previous papers have been embedding graphs low-dimensional hyperbolic space [7] setting, which yields better performance than such of Euclidean space. However, in the case of high-dimensional space, empirical results presented in (Frogner et al., 2019) are in line with Gromov Hyperbolicity discussed in section 2.5.

---

[7] $d$-dimensional Poincaré ball, $d \in (2, \ldots 14)$.

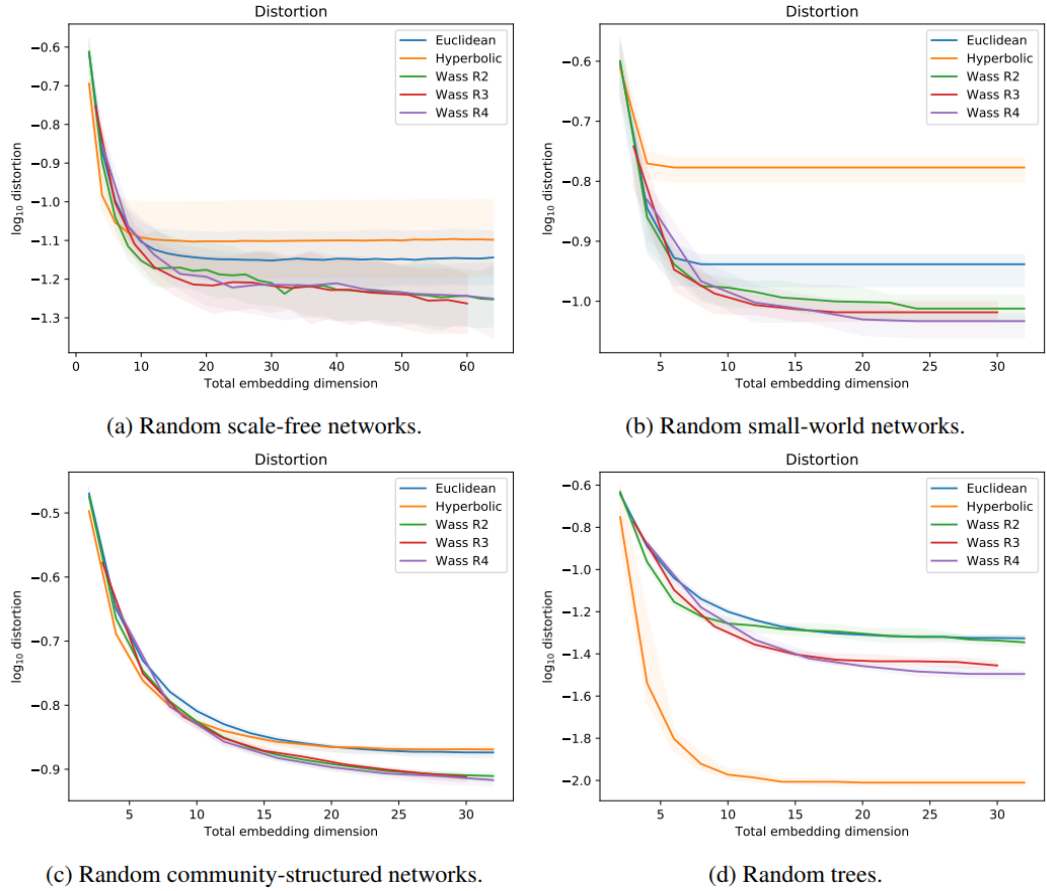

**Fig. 3.2.:** Comparison of distortion of various random networks embeddings in different spaces; reproduced from Frogner et al. (2019)

Therefore, we can see that in low dimensional case, work by Frogner et al. (2019) agrees with previous papers, while, in the high dimensional case, results are in line with Gromov Hyperbolicity theory. We find this observation very interesting as it will come useful in chapter 6, when discussing the performance of the Hyperbolic VGG and GCN networks.

## 3.2 Hyperbolic Neural Networks

In the final section of this chapter, we would like to review papers which serve as the primary basis for this work. As we have seen in the previous sections, researchers have previously used hyperbolic spaces as embedding spaces in machine learning algorithms; however, only in shallow models setting. An arbitrary neural network can be seen as a sequence of regular Euclidean operations $f : \mathbb{R}^n \to \mathbb{R}^m$. Therefore, making use of hyperbolic spaces in deep models would mean changing those to hyperbolic operations $f : \mathbb{D}^n \to \mathbb{D}^m$ such that not just the final embedding space but the latent space at each layer is hyperbolic. "Hyperbolic Neural Networks" paper by Ganea et al. (2018a) attempts at precisely that, making each layer of neural network map data to Poincaré disk.

Generally, we can regard multilayer perceptron as the most fundamental class of feedforward Neural Networks. If we could adapt it to the hyperbolic case, we could further adapt any class of feedforward neural networks, including CNNs and GCNs, since they can be treated as a specific case of a multilayer perceptron (see section 4.1) with additional operations. However, before going into how we can do that, it is crucial to review building blocks of a multilayer perceptron model.

The simplest multilayer perceptron consists of three successive layers: a linear input layer, hidden linear layer, and output linear layer (see fig. 3.3) with activation function between each of the layers. Task-depending classifier can be used on top of the output layer to make a prediction. Since we are interested in doing a classification task in this work, we will use softmax function as the classifier. In the case of softmax, the class with the highest probability is the predicted one, and the model is trained using cross-entropy loss with a one-hot vector of the actual class.

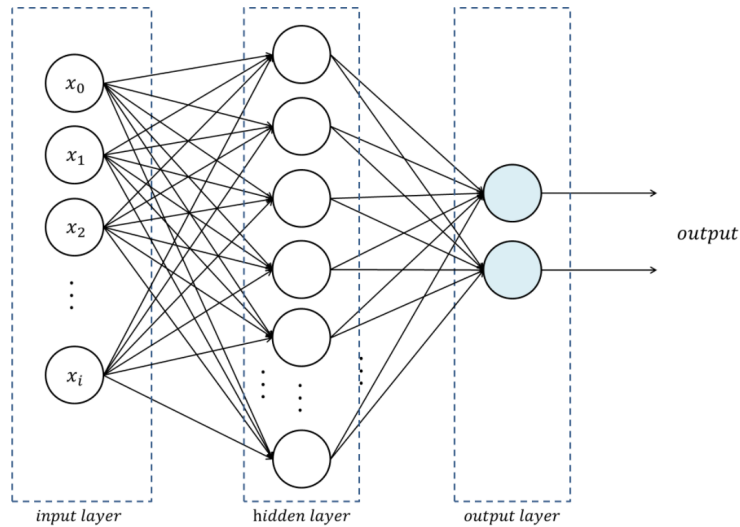

**Fig. 3.3.:** MLP visualization; reproduced from Sungtae An.

Therefore, based on the description of the Multilayer perceptron, we need to redefine three building blocks to make it fully hyperbolic:

1. Linear Layer: $x^{(l+1)} = W^{(l)}\mathbf{x}^{(l)} + \mathbf{b}^{(l)}$, where $l$ is the layer index, $W$ is the weight matrix, $\mathbf{x}$ is the output of the previous layer and $\mathbf{b}$ is the bias vector.
2. Pointwise nonlinearity: $\phi(\mathbf{x}) = (\phi(x_1), \ldots \phi(x_n))$
3. Softmax Layer: $p(y = k|\mathbf{x}) = \frac{e^{x_k}}{\sum_{i=1}^{n} e^{x_i}}$

It is already possible to define hyperbolic softmax.

> **Theorem 5: Hyperbolic softmax (Ganea et al., 2018a)**
>
> Given $K$ classes and $k \in \{1, \ldots K\}$, $\mathbf{p}_k \in \mathbb{D}_c^n, \mathbf{a}_k \in T_{\mathbf{p}_k}\mathbb{D}_c^n \setminus \{\mathbf{0}\}$, hyperbolic softmax can be defined as following:
>
> $$p(y=k|\mathbf{x}) \propto \exp\left(\frac{\lambda_{\mathbf{p}_k}^c \|\mathbf{a}_k\|}{\sqrt{c}} d_c(x, \tilde{H}_{a,p}^c))\right), \qquad \forall \mathbf{x} \in \mathbb{D}_c^n$$
>
> $$d_c(\mathbf{x}, \tilde{H}_{\mathbf{a},\mathbf{p}}^c)) := \inf_{w \in \tilde{H}_{\mathbf{a},\mathbf{p}}^c} d_c(\mathbf{x}, \mathbf{w})$$
>
> $$= \sinh^{-1}\left(\frac{2\sqrt{c}((-\mathbf{p}_k \oplus_c \mathbf{x}) \cdot \mathbf{a}_k)}{(1 - c\|-\mathbf{p}_k \oplus_c \mathbf{x}\|^2)\|\mathbf{a}_k\|}\right)$$
>
> where $\tilde{H}_{a,p}^c$ is a hyperplane in a Poincaré disk.
>
> Since $\mathbf{a}_k \in T_{\mathbf{p}_k}\mathbb{D}_c^n$ and thus depends on $\mathbf{p_k}$, we separate optimization of $\mathbf{a}_k$ from the optimization of $\mathbf{p}_k$. We do so by introducing new variable $\mathbf{a}_k' \in T_\mathbf{0}\mathbb{D}_c^n = \mathbb{R}^n$ which we optimize separately and then parallel-transport to the tangent plane of $\mathbf{p}_k$. Mathematically, it can be written as following:
>
> $$\mathbf{a}_k = P_{\mathbf{0} \to \mathbf{p}_k}^c(\mathbf{a}_k') = \frac{\lambda_\mathbf{0}^c}{\lambda_\mathbf{p}^c}\mathbf{a}_k', \qquad \mathbf{a}_k' \in T_\mathbf{0}\mathbb{D}_c^n = \mathbb{R}^n$$

> **Note 2**
>
> Since the definition is essentially a combination of Hyperbolic Linear Layer and Softmax, as shown in (Ganea et al., 2018a), we will call it **HypMLR** for hyperbolic multinomial logistic regression.

In the theorem above, hyperplane $\tilde{H}_{a,p}^c$ can be seen as the union of images of all geodesics in $\mathbb{D}_c^n$ orthogonal to $\mathbf{a}$ and containing $\mathbf{p}$.

Visualization of the difference between hyperbolic and Euclidean softmax hyperplanes is presented in fig. 3.4.

To define next two operations, it is important to recollect that, as discussed in (4), Möbius scalar multiplication and Gyroline can be rewritten as a sequence of logarithmic map, scalar multiplication and exponential map i.e. $r \otimes_c \mathbf{u} = \exp_\mathbf{0}^c(r \log_\mathbf{0}^c(\mathbf{u}))$. We will further call such approach **log-exp-map treatment**.

Using such interpretation, authors suggest Möbius version of Euclidean operations. It suggests a universal form for the definitions of the rest of MLP building blocks for the hyperbolic space with a few exceptions.

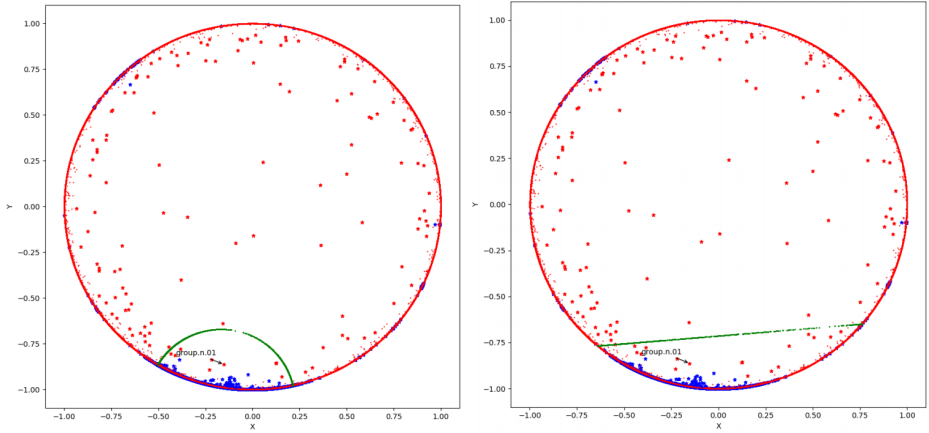

**Fig. 3.4.:** Hyperplane learned by hyperbolic softmax (left) and Euclidean softmax (right); reproduced from Ganea et al. (2018a).

> **Definition 23: Möbius version (Ganea et al., 2018a)**
>
> For $f : \mathbb{R}^n \to \mathbb{R}^m$, Möbius version of $f$ as $f^{\otimes_c} : \mathbb{D}_c^n \to \mathbb{D}_c^m$:
>
> $$f^{\otimes_c}(\mathbf{x}) = \exp_{\mathbf{0}}^c \left( f \left( \log_{\mathbf{0}}^c (\mathbf{x}) \right) \right)$$
>
> $f^{\otimes_c}$ satisfies following properties:
> - Homomorphism condition: $(f \circ g)^{\otimes_c}(\mathbf{x}) = f^{\otimes_c} \circ g^{\otimes_c}(\mathbf{x}), f : \mathbb{R}^n \to \mathbb{R}^m, g : \mathbb{R}^l \to \mathbb{R}^m$
> - Direction preserving: $\frac{f^{\otimes_c}(\mathbf{x})}{\|f^{\otimes_c}(\mathbf{x})\|} = \frac{f(\mathbf{x})}{\|f(\mathbf{x})\|} \ \forall f(\mathbf{x}) \neq \mathbf{0}$

> **Note**
>
> If we take $f : \mathbb{R}^n \to \mathbb{R}^n$ to be a pointwise nonlinearity, then $f^{\otimes_c}$ can be seen as hyperbolic version of a nonlinear activation function.

Now we need to adapt the most crucial part of MLP - linear layer. Intuitively, we could adapt it by merely doing a Möbius version of the linear layer (see fig. 3.5a). The main problem with this approach, however, is that it would be equivalent to merely doing all operation in Euclidean space and then mapping back to the hyperbolic space in the very end to perform hyperbolic softmax (see fig. 3.5b). Such a sequence of operations, though, would not lead to any improvements since our model will stay inherently Euclidean. Therefore, we are forced to enrich our building blocks with operations which are entirely hyperbolic (see fig. 3.5c), meaning that all the operands already exist in the hyperbolic space before the operation is performed.

Ganea et al. (2018a) suggest to solve this issue by splitting Linear Layer into two operations, namely, matrix multiplication and bias translation. Since bias translation can

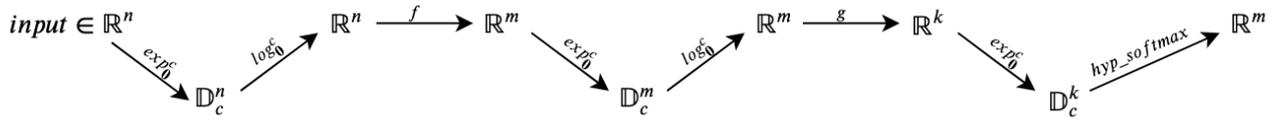

(a) Full mappings between Hyperbolic and Euclidean spaces.

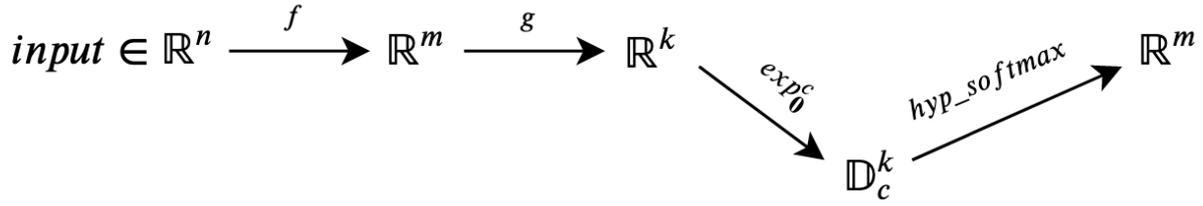

(b) Simplified mappings between Hyperbolic and Euclidean spaces.

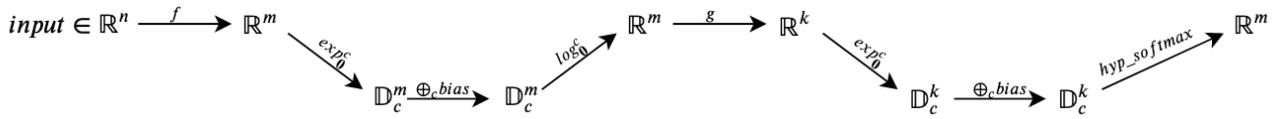

(c) Mappings between Hyperbolic and Euclidean spaces with added bias translation operation.

**Fig. 3.5.:** Comparison of models with and without fully hyperbolic operations.

be interpreted as vector addition in the hyperbolic space, it is equivalent to gyrovector addition.

Definition provided in (23) allows for a straightforward definition of Möbius matrix-vector multiplication.

> **Definition 24: Möbius matrix-vector multiplication (Ganea et al., 2018a)**
>
> If $\mathbf{M} : \mathbb{R}^n \to \mathbb{R}^m$, is a linear map, which can be represented in a matrix form, then $\forall \mathbf{x} \in \mathbb{D}_c^n$:
>
> $$\mathbf{M}^{\otimes_c}(\mathbf{x}) = \mathbf{0} \qquad \text{if } \mathbf{M}\log_0^c(\mathbf{x}) = \mathbf{0}$$
>
> $$\mathbf{M}^{\otimes_c}(\mathbf{x}) = \exp_0^c(\mathbf{M}\log_0^c(\mathbf{x}))$$
> $$= \frac{1}{\sqrt{c}} \tanh\left(\frac{\|\mathbf{M}\mathbf{x}\|}{\|\mathbf{x}\|} \tanh^{-1}(\sqrt{c}\|\mathbf{x}\|)\right) \frac{\mathbf{M}\mathbf{x}}{\|\mathbf{M}\mathbf{x}\|} \qquad \text{otherwise}$$
>
> If we write $\mathbf{M}^{\otimes_c}(\mathbf{x})$ in a more familiar form $\mathbf{M} \otimes_c \mathbf{x}$, where $\mathbf{M} \in \mathbb{R}^{m \times n}, \mathbf{x} \in \mathbb{D}_c^n$, it has following properties:
> - Matrix associativity: $(\mathbf{M}\mathbf{M}') \otimes_c \mathbf{x} = \mathbf{M} \otimes_c (\mathbf{M}' \otimes_c \mathbf{x})$, where $\mathbf{M} \in \mathbb{R}^{m \times n}, \mathbf{M}' \in \mathbb{R}^{n \times l}$
> - Scalar-matrix associativity: $(r\mathbf{M}) \otimes_c \mathbf{x} = r \otimes_c (\mathbf{M} \otimes_c \mathbf{x})$, where $r \in \mathbb{R}, \mathbf{M} \in \mathbb{R}^{m \times n}$
> - Rotation preservation: $\mathbf{M} \otimes_c \mathbf{x} = \mathbf{M}\mathbf{x} \qquad \forall \mathbf{M} \in \mathcal{O}_n(\mathbb{R})$

Lastly, the definition of bias translation using Möbius vector addition can be directly derived using parallel transport.

> **Definition 25: Bias translation (Ganea et al., 2018a)**
>
> Möbius translation of a point $\mathbf{x} \in \mathbb{D}_c^n$ by a bias $\mathbf{b} \in \mathbb{D}_c^n$ is given by:
>
> $$\mathbf{x} \oplus_c \mathbf{b} = \exp_\mathbf{x}^c(P_{0 \to \mathbf{x}}^c(\log_0^c(\mathbf{b}))) = \exp_\mathbf{x}^c\left(\frac{\lambda_0^c}{\lambda_\mathbf{x}^c} \log_0^c(\mathbf{b})\right)$$

Note that, similarly to gyrovector operations defined in section 2.4, in limit $c \to 0$, all operations recover their Euclidean form.

Finally, all necessary blocks for building MLPs have been defined. There are two essential things to notice about the suggested operations. First of all, they can be implemented efficiently with any modern deep learning framework since (a) all parameters are being optimized in the Euclidean space, so no new optimization algorithms are necessary; (b) Möbius matrix-vector multiplication can be processed on a GPU since matrix multiplication itself does not change which significantly speeds up the performance. Secondly, given the functional form of Möbius matrix-vector multiplication allows us to generalize it to the case of convolutional neural networks, which we will demonstrate in the next chapter.

> **Note 3**
>
> In the rest of this work, to accentuate that layer is a Möbius version of more common Euclidean neural network layer, we will add prefix 'Hyp' to the name of the layer, e.g., HypLinear, HypBias, HypDropout.

Although being the most comprehensive in its attempt to generalize neural networks to the hyperbolic case, the paper by Ganea et al. (2018a) is not the only paper that tried to see whether modern neural architectures can benefit from using hyperbolic space as the embedding space. The other important paper is "Hyperbolic Attention Networks" by Gulcehre et al. (2018). Unlike "Hyperbolic Neural Networks" paper, it does not attempt at making all layers hyperbolic but, instead, focuses on seeing whether Attention-based models can benefit from hyperbolic attention.

Since attentive read operation lies at the core of attention computational module, if we could generalize it to hyperbolic space, it would allow for further generalization of arbitrary attention modules.

> **Definition 26: Attentive Read (Gulcehre et al., 2018)**
>
> Attentive read operation is defined as:
>
> $$\mathbf{r}(\mathbf{q}_i, \{\mathbf{k}_j\}_j) = \frac{1}{Z} \sum_j \alpha(\mathbf{q}_i, \mathbf{k}_j) \mathbf{v}_{ij}$$
>
> where $\alpha(\cdot, \cdot) \mapsto \alpha \in \mathbb{R}$ computes a scalar matching form, $\mathbf{q}_i$ is a vector called the query, the $\mathbf{k}_j$ are *keys* for the memory locations being read from, the vector $\mathbf{v}_{ij}$ is a value to be read from location $j$ by query $i$, $Z > 0$ is a normalization factor. Attentive read operation can be split into two parts:
> 1. Matching: compute attention weights $\alpha_{ij}$
> 2. Aggregation: compute $\mathbf{r}(\mathbf{q}_i, \{\mathbf{k}_j\}_j)$ using attention weights

Authors use split into matching and aggregation to suggest hyperbolic attention.

1. **Hyperbolic matching**: given $\mathbf{q}_i$ and $\mathbf{k}_j$, mapped into a hyperbolic space, attention weights $\alpha_{ij}$ are calculated as $\alpha(\mathbf{q}_i, \mathbf{k}_j) = f(-\beta d_{\mathbb{H}}(\mathbf{q}_i, \mathbf{k}_j) - c)$, where $d_{\mathbb{H}}$ is the distance in hyperboloid model of hyperbolic space and $\beta$ and $c$ are parameters. Depending on the form of $f$, appropriate normalization can be chosen.

2. **Hyperbolic aggregation**: the generalization of attention to the hyperbolic case is straightforward once we look at the definition of Möbius weighted midpoint ((20)). Authors use its Einstein version[8] to define aggregation operation as following: $m_i(\{\alpha_{ij}\}_j, \{\mathbf{v}_{ij}\}_j) = \sum_j \left[ \frac{\alpha_{ij} \gamma(\mathbf{v}_{ij})}{\sum_l \alpha_{il} \gamma(\mathbf{v}_{ij})} \right] \mathbf{v}_{ij}$, where $\gamma(\mathbf{v}_{ij})$ are the Lorentz factors.

---
[8] It corresponds to gyromidpoint in Klein model of hyperbolic space.

Hyperbolic attention can be summarized in a computational graph, see fig. 3.6.

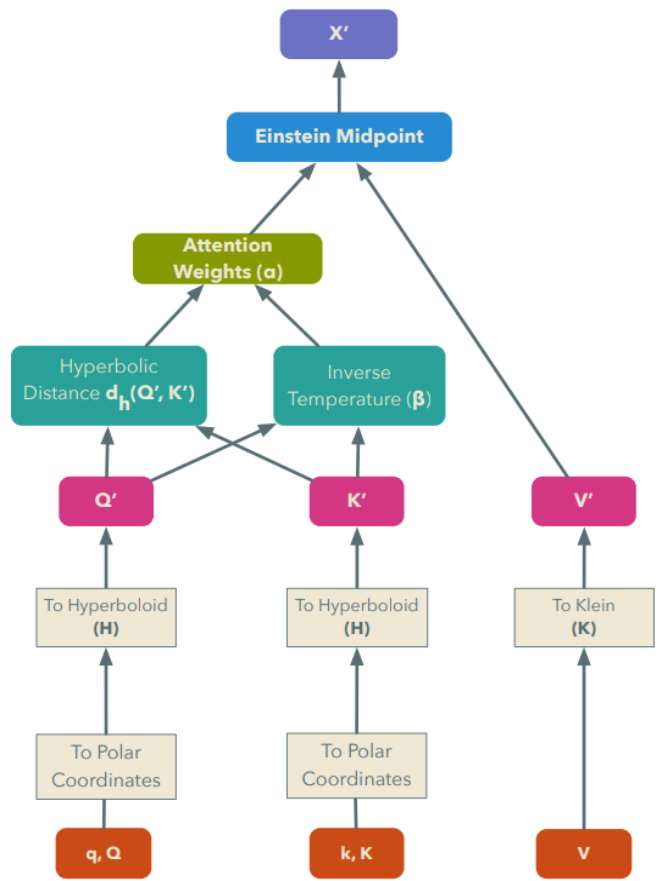

**Fig. 3.6.:** Computational graph of hyperbolic attention operation; reproduced from Gulcehre et al. (2018).

Although not directly relevant to our goal, this interpretation of attention will come very valuable in further chapters. Additionally, authors show that common attention-based architectures (Vaswani et al., 2017; Santoro et al., 2017) benefit from making attention modules hyperbolic. It is demonstrated on the tasks of neural machine translation and VQA.

We conclude the chapter with a discussion of one of the most recent papers on using hyperbolic spaces in the context of deep learning, namely "Hierarchical Representations with Poincaré Variational Auto-Encoders" by Mathieu et al. (2019).

In this paper, authors try to learn latent space defined over hyperbolic space. From the perspective of learning latent space defined on spaces with constant negative curvature, this work can be seen as a follow-up work for "Hyperspherical Autoencoders" by Davidson et al. (2018).

Essentially, the paper has three main points, namely, discussion of encoder, decoder, and latent space distribution. The encoder is defined as a simple feedforward neural network. Since the output of the encoder is a parametrization of wrapped normal distribution on hyperbolic spaces (Nagano et al., 2019), no additional architectural changes are necessary compared to the original VAE encoder suggested in (Kingma and Welling, 2013).

Next, reparametrisation is done using hyperbolic polar coordinate. Given outputs $\mathbf{v}$ and $\boldsymbol{\sigma}$ of the encoder, sample $\mathbf{x} \sim \mathcal{N}_{\mathbb{D}_c^n}(\cdot|\exp_0^c(\mathbf{v}), \sigma^2)$ is generated as:

$$\mathbf{x} = \exp_\mu^c(\frac{r}{\lambda_\mu^c}\boldsymbol{\alpha}),$$

$$\boldsymbol{\alpha} \sim \mathcal{U}(\mathbb{S}^{d-1}) \quad \text{where } \mathbb{S}^{d-1} \text{ is } d-1 \text{ hypersphere,}$$

$$\text{PDF}(r) = \frac{\mathbb{1}_{\mathbb{R}_+}(r)}{Z_r(\sigma)} e^{-\frac{r^2}{2\sigma^2}} \left(\frac{\sinh(\sqrt{c}r)}{\sqrt{c}}\right)^{d-1} \quad \text{where } Z_r(\sigma) \text{ is normalizing constant}$$

$\boldsymbol{\alpha}$ is the direction of hyperbolic polar coordinate while $r$ is the hyperbolic radius. Note that $r$ can not be sampled directly due to the absence of closed-form CDF. Therefore, rejection sampling is used to sample $r$.

Finally, samples are fed into the decoder. The decoder uses a concatenation of hyperbolic softmax operations, defined in (5), as the first layer mapping from hyperbolic space to Euclidean space. Afterwards, samples are processed through a generic neural network.

Authors show that this yields superior results compared to standard VAE when applied to a synthetic dataset created from a branching diffusion process. Dataset consists of points $\mathbf{x}_i \in \mathbb{R}^n$, which are sampled from the following sampling process:

$$\mathbf{x}_i \sim \mathcal{N}(\cdot|\mathbf{x}_{\pi(i)}, \sigma_0^2) \qquad \forall i \in 1, \ldots N$$

where $\pi(i)$ is the index of the $i$-th node parent. For each parent, $b$ nodes are sampled with $b$ being the branching factor. Nodes are sampled until maximum depth $d$ is reached. Node $x_0$ has a depth of 0. Therefore, generated data has a known hierarchical structure. However, the model only has access to the vector representations and not structure itself.

Finally, noisy samples are generated around each node $\mathbf{x}_i$ through

$$\mathbf{y}_{i,j} = \mathbf{x}_i + \epsilon_{i,j}, \qquad \epsilon_{i,j} \sim \mathcal{N}(\cdot|\mathbf{0}, \sigma_j^2), \qquad \forall i \in 1, \ldots N, \forall j \in 1, \ldots J$$

Authors choose vector dimensionality $n$ to be 50 and $J$ to be 5. $\mathbf{x}_0 = \mathbf{0}, \sigma_0 = 1, \sigma_j = \frac{1}{5}$. Branching factor is set to $b = 2$ and maximum depth is $d = 6$.

Results of using Poincaré VAE on this dataset can be seen in fig. 3.7.

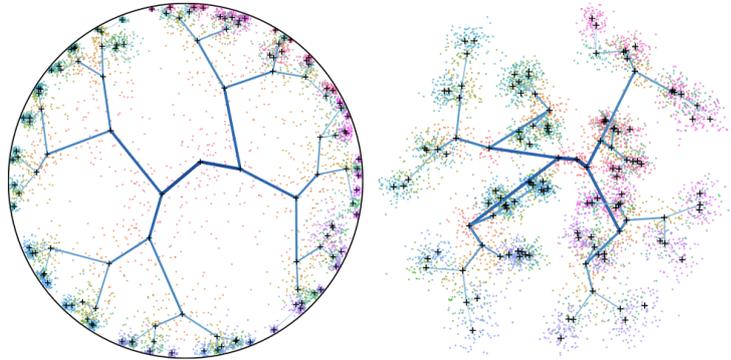

**Fig. 3.7.:** Embedding of branching diffusion process using Poincaré VAE (left) and standard VAE (right); reproduced from Mathieu et al. (2019).

The suggested way of creating hierarchical dataset will come extremely useful in the later chapters since we have taken it as an inspiration for creating tree-like graph dataset.

As we have seen in this chapter, hyperbolic spaces have already proven their usefulness for the tasks that work with data that has a hierarchical structure. This consideration serves as a valuable motivation for generalizing existing feedforward architectures to work on image and graph data. With this concludes current chapter and now we will proceed to the description of our approach.

# 4

# Hyperbolic Convolutional Neural Networks

In this chapter, we introduce a generalization of the convolutional neural networks and graph convolutional networks for the hyperbolic space. Also, since modern neural network models use additional layers for improved performance, we will generalize the most common blocks to the hyperbolic case. These will be used in the next chapter for making a comparison between identical Euclidean and Hyperbolic models. We start by a discussion of Euclidean versions of CNNs and GCNs and then proceed to define their hyperbolic versions.

So far, we have discussed existing approaches for making MLPs hyperbolic. While vanilla neural networks work with Euclidean feature vectors $\mathbf{x} \in \mathbb{R}^n$ at each layer, Hyperbolic Neural Networks suggests using hyperbolic feature vectors $\mathbf{x} \in \mathbb{D}^n$. However, the main drawback of suggested models is that there is not a straightforward way to use them in the context of convolutional neural networks.

It is crucial to realize that convolutional neural networks can be seen as MLPs that operate on fields of feature vectors on $\mathbb{R}^2$ in case of images, where $\mathbb{R}^2$ is the space of pixel coordinates or $\mathscr{V}$, where $\mathscr{V}$ is a set of vertices in case of graphs:

$$f : \mathbb{R}^2 \to \mathbb{R}^{c_{in}} \quad and \quad f : \mathscr{V} \to \mathbb{R}^n$$

Therefore, our goal is to have hyperbolic convolutional layers operating on fields of feature gyrovectors:

$$f : \mathbb{R}^2 \to \mathbb{D}_c^{c_{in}} \quad and \quad f : \mathscr{V} \to \mathbb{D}_c^n$$

Importantly, note that in both of the cases, we do not require base space $\mathbb{R}^2$ or $\mathscr{V}$, on which field is defined, to change since we only care about changing the geometry of feature vector spaces and not the base space itself.

## 4.1 Convolutional Neural Networks

We want to start off by providing a mathematical interpretation of the convolutional neural networks that are necessary for understanding our work. In doing so, we assume that the reader is familiar with the basics of convolutional neural networks.

First, it is important to introduce a view of images and stacks of feature map as functions. We model images as being supported on a bounded, typically square and discrete, domain, which we regard as base space. In this case, image takes following functional form: $f : \mathbb{R}^2 \to \mathbb{R}^{c_{in}}$, where $c_{in}$ is the number of channels[9]. Essentially, function $f$ can be seen as mapping from each pixel coordinate $(i, j)$ to a certain feature vector $f(i, j) \in \mathbb{R}^{c_{in}}$. Note that by changing $\mathbb{R}^2$ to a different space (e.g., $\mathbb{R}^3$ for 3D-data or $\mathcal{V}$ for graphs) we allow for similar interpretation for data coming from a non-2D domain.

So, to sum up, image can be seen as a feature vector field $f : \mathbb{R}^2 \to \mathbb{R}^{c_{in}}$ with feature vectors $\mathbb{R}^{c_{in}}$ being attached to every point in a base space $\mathbb{R}^2$, see fig. 4.1.

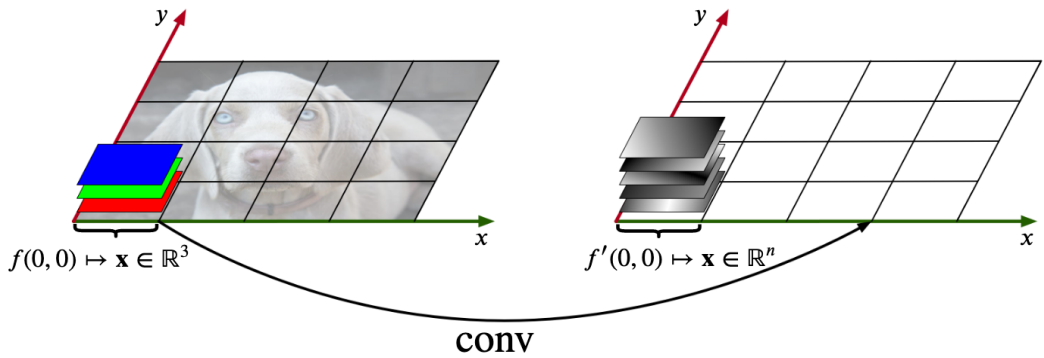

**Fig. 4.1.:** Visualization of images as feature vector fields. Convolution operation is transforming feature vectors attached to each point of a field.

This allows for the following interpretation of a convolution operation:

$$g(x, y) = \sum_{s=-a}^{a} \sum_{t=-b}^{b} w(s, t) f(x - s, y - t)$$

where $x, y \in \mathbb{R}$, $g : \mathbb{R}^2 \to \mathbb{R}^{c_{out}}$ is the filtered image, $g(x, y) \in R^{c_{out}}$ is a feature vector of filtered image at pixel with coordinates $(x, y)$, $f : \mathbb{R}^2 \to \mathbb{R}^{c_{in}}$ is the original image, $f(x, y) \in \mathbb{R}^{c_{in}}$ is a feature vector of original image at pixel with coordinates $(x, y)$, $w \in \mathbb{R}^{k_{width} \times k_{height} \times c_{in} \times c_{out}}$ is the filter kernel with kernel width $k_{width}$ and height $k_{height}$ and $w(x, y) \in \mathbb{R}^{c_{in} \times c_{out}}$ being the matrix for performing linear transformation on feature vector corresponding to filter kernel coordinate $(x, y)$.

---

[9]Number of channels is equal to 3 in case of RGB-images or 1 in case of grayscale.

$$input \in \{f_0 | f_0 : \mathbb{R}^2 \to \mathbb{R}^{c_{in}}\} \xrightarrow{Conv2D} \{f_1 | f_1 : \mathbb{R}^2 \to \mathbb{R}^{c_1}\} \xrightarrow{Conv2D} \{f_2 | f_2 : \mathbb{R}^2 \to \mathbb{R}^{c_2}\} \xrightarrow{pool} \mathbb{R}^n$$

**(a)** Typical mappings in CNNs.

$$input \in \mathbb{R}^n \xrightarrow{f} \mathbb{R}^{c_1} \xrightarrow{g} \mathbb{R}^{c_2}$$

**(b)** Typical mappings in MLPs.

**Fig. 4.2.:** Mathematical view of mappings between spaces in CNNs and MLPs.

## 4.2 Batch Normalization

Before we proceed to discuss graph convolutional neural networks, we would like to mention an essential technique for improving the performance of CNNs, namely Batch Normalization (Ioffe and Szegedy, 2015). Since it allows for more efficient training of modern deep convolutional networks, deriving its hyperbolic version is of paramount importance.

Batch normalization is a technique dedicated to fighting covariate shift (Shimodaira, 2000), which is caused by changes in the input distribution of the learning system. Although the original paper discussed learning system as a whole, Ioffe and Szegedy (2015) suggest that it is possible to see the deep neural network as a set of learning systems, with each layer being a separate learning system. Such interpretation allows for defining internal covariance shift - change in the distribution of network activations, caused by changes in network parameters during training. Since inputs to each layer of a deep neural network are amplified by the parameters of all the preceding layers, internal covariate shift results in that layers need to continuously adapt to the new distribution. Such adaptation is very harmful to the performance of deep neural networks.

Whitening can be seen as one of the previous steps in the direction of avoiding internal covariate shift and has been shown (Wiesler and Ney, 2011; LeCun et al., 2012) to result in faster convergence of neural networks. However, in the batch setting, using the whole set for normalizing activations yields impractical and computationally expensive. Also, whitening each layer's input is not everywhere differentiable and, again, computationally expensive. Therefore, two simplifications have to be made:

- Each channel scalar feature is normalized independently such that it has mean of zero and variance of one;

- Estimates of the mean and variance of each activation are produced per mini-batch and spatial dimensions.

Since this might harm the representational power of the network, authors of (Ioffe and Szegedy, 2015) suggest that this problem can be alleviated by changing transformation inserted in the network such that it can represent the identity transform.

All these ideas together result in the following algorithm:

---
**Algorithm 1** Batch Normalization (Ioffe and Szegedy, 2015)

---
**Input:** Value of $x$ over a mini-batch $\mathbb{B} = \{x_1 \ldots x_m\}$

**Input:** Parameters to be learned $\gamma, \beta$

**Output:** $\{y_i = BN_{\gamma,\beta}(x_i)\}$

1: $\mu_{\mathbb{B}} \leftarrow \frac{1}{m} \sum_{i=1}^{m} x_i$ ▷ Mini-batch Mean
2: $\sigma_{\mathbb{B}}^2 \leftarrow \frac{1}{m} \sum_{i=1}^{m} (x_i - \mu_{\mathbb{B}})^2$ ▷ Mini-batch Variance
3: $\hat{x}_i \leftarrow \frac{x_i - \mu_{\mathbb{B}}}{\sqrt{\sigma_{\mathbb{B}}^2 + \epsilon}}$ ▷ Normalize
4: $y_i \leftarrow \gamma \hat{x}_i + \beta \equiv BN_{\gamma,\beta}(x_i)$ ▷ Scale and shift

---

Batch Normalization has been shown to significantly improve the performance of convolutional neural networks while also speeding up the training. We will come back to Batch Normalization in the next chapters to show how it can be generalized to act in the hyperbolic space. We will also review VGG architecture in more detail in chapter 4.

## 4.3 Graph Convolutional Neural Networks

In the previous section, we have been discussing CNNs as a conventional algorithm for working with images. However, images have a lot in common with graphs since they can be seen as grids. Therefore, the idea of defining hyperbolic versions of graph convolutional neural networks arises almost naturally.

There have been various approaches to working with graphs (Duvenaud et al., 2015; Kipf and Welling, 2016; Li et al., 2015; Jain et al., 2015; Bruna et al., 2013; Henaff et al., 2015), however, we will focus on graph convolutional networks as suggested in (Kipf and Welling, 2016) being the most universal and widely used architecture.

Graph convolutional networks work with graph $\mathcal{G} = (\mathcal{V}, \mathcal{E})$. Mathematical definition of graphs in that case is equivalent to such of images, replacing $\mathbb{R}^2$ with $\mathcal{V}$ yielding $f : \mathcal{V} \to \mathbb{R}^n$ as a functional form for graphs. In this case $f$ assings a feature vector $f(v)$ to each node $v \in \mathcal{V}$.

The main goal of a graph convolutional network is to extract node feature vectors $f(v)$ for the final downstream task, such as classification or link prediction. Graph convolutional networks are often deep with multiple layers stacked on top of each other with nonlinearities in-between. Each layer is parameterized by:

- Node feature $x_i$ for every node $i$, or, in matrix form, $X \in \mathbb{R}^{n\_nodes \times n}$, where $n\_nodes$ is the number of nodes in a graph;
- Graph structure description in a form of an adjacency matrix $A \in \mathbb{R}^{n\_nodes \times n\_nodes}$, where $n\_nodes$ is the number of nodes in a graph;
- Learnable weight matrix $W \in \mathbb{R}^{n \times m}$.

In this case, the update function of every layer can be written as the following function:

$$X^{(l+1)} = g(f(X^{(l)}, A, \mathbf{W}^{(l)})),$$

where $l$ is the layer order, $f$ is the graph processing function and $g$ is a chosen activation function.

Kipf and Welling (2016) suggest to write update function as following:

$$\begin{aligned} X^{(l+1)} &= g(D^{-\frac{1}{2}}(A+I)D^{-\frac{1}{2}}X^{(l)}W^{(l)}) \\ &= g(A'X^{(l)}\mathbf{W}^{(l)}), \end{aligned}$$

with $D$ being the diagonal node degree matrix of $(A+I)$.

Layer using such update function is called GCNConv.

This model solves two problems that were encountered with old approaches.

First of all, it adds self-loops to the adjacency matrix [10], which means that when doing the propagation step, we sum up not only the feature vectors of the neighbouring nodes but also of the node itself.

Secondly, it changes normalization with $D^{-1}A$, which results in mere averaging of the neighboring nodes, by symmetric normalization $D^{-\frac{1}{2}}(A+I)D^{-\frac{1}{2}}$, which yields more interesting dynamics. This idea of choosing better normalization coefficients was further developed in Graph Attention Networks (Veličković et al., 2017), in which authors suggest to learn normalization coefficients using multi-head attention.

---

[10]This is done through $(A+I)$.

The pure intuition behind this approach is that its neighbours can characterize each node in a graph. However, using standard CNNs similarly to image processing yields impossible since there is no natural ordering of the nodes of the graph. Therefore, all nodes share the same weight matrix hence making such network convolutional.

Another critical point to take note of is that if we stack multiple GCNConv layers, we mainly propagate signal from neighbourhoods which are further than distance 1 away from the current node, see fig. 4.3. It allows us to capture dependencies between arbitrarily distant nodes, which is analogous to receptive field growth in deep CNNs.

More general visualization with additional sampling step, as suggested in (Hamilton et al., 2017), can be seen in fig. 4.3.

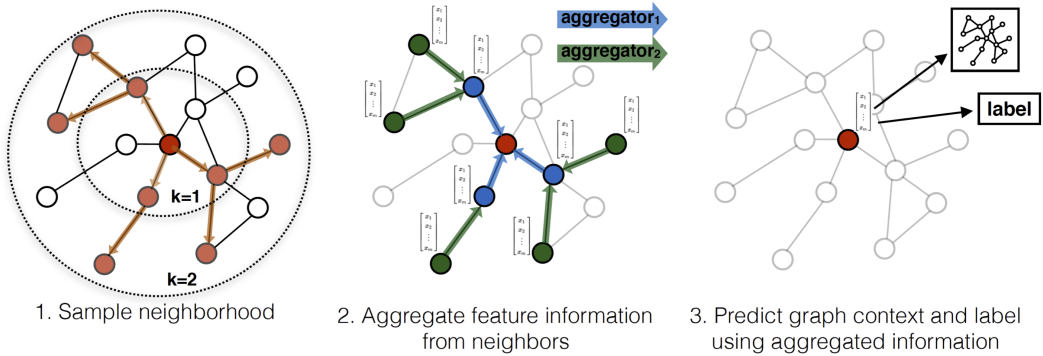

**Fig. 4.3.:** Graph Convolutional Layer Propagation, as defined in GraphSage (Hamilton et al., 2017).

Interestingly enough, such a GCNConv layer has an interpretation in terms of a more well-known message passing algorithm (Gilmer et al., 2017; Yedidia, 2011).

In the most general case, message passing on graphs takes on the following the form:

$$m_v^{t+1} = \sum_{w \in N(v)} M_t(h_v^t, h_w^t, c_{vw})$$
$$h_v^{t+1} = U_t(h_v^t, m_v^{t+1}),$$

where $N(v)$ is the neighbourhood of vertex $v$, $m_v^{t+1} \in \mathbb{R}^n$ is constructed by aggregating messages passed to the vertex $v$ at timestep $t+1$, $h_v^{t+1} \in \mathbb{R}^m$ is the feature vector of vertex $v$ at timestep $t+1$, $M_t : \mathbb{R}^n \times \mathbb{R}^n \times \mathbb{R} \to \mathbb{R}^m$ is the message function, $c_{vw} \in \mathbb{R}$ is message weight coefficients and $U_t : \mathbb{R}^n \times \mathbb{R}^m \to \mathbb{R}^m$ is the update function at timestep $t$.

It gets more intuitive if we take a look at how the GCNConv layer discussed above translates to message-passing terms.

$$M_t(h_v^t, h_w^t, c_{vw}, \mathbf{W}^t) = c_{vw} \mathbf{W}^t h_w^t$$
$$c_{vw} = (deg(v)deg(w))^{-\frac{1}{2}} A_{vw}$$
$$U_v^t(h_v^t, m_v^{t+1}) = g(m_v^{t+1})$$

where deg($v$) is the node degree of vertex $v$. Note that $c_{vw}$ still amounts for the same symmetric normalized Laplacian $D^{-\frac{1}{2}}(A+I)D^{-\frac{1}{2}}$ if stacked into a matrix form.

So, let us analyze what graph message passing for GCNConv layer does at every timestep. First, all the messages are computed by multiplying feature vectors of each node by a weight matrix. Next, they are multiplied by a corresponding normalization coefficient and passed over the edges to the neighbouring nodes. Each node sums up all the incoming messages and applies a chosen nonlinearity. The output is used as an updated feature vector for the corresponding node.

This interpretation yields crucial for many applications (Wang et al., 2018; Yoon et al., 2018; Garcia Satorras et al., 2019) and, when combined with Graph Attention Networks (Veličković et al., 2017), allows us to come up with a generalization of graph convolutional networks to the hyperbolic case, which we will discuss in more detail in section 4.5.

## 4.4 Hyperbolic Convolutional Neural Network

Finally, we are ready to suggest a way to generalize the most popular layers used in the modern deep convolutional neural networks to make them hyperbolic. We will start with the most essential, namely convolutional, layer.

### 4.4.1 Hyperbolic Convolutional Layer

The key to a generalization of a convolutional layer to the hyperbolic space lies in the interpretation of convolution as an operation over feature vector field. We have previously discussed that image is essentially feature vector field $f : \mathbb{R}^2 \to \mathbb{R}^n$ with feature vectors $\mathbb{R}^n$ attached to every point in a base space $\mathbb{R}^2$.

Now, if we would have a single feature vector, we would approach it with a simple linear layer and could make use of suggested Hyperbolic Linear Layer (24). This operation amounts to taking a feature vector (denoted as $\mathbf{x} \in \mathbb{D}^n$ for the definiteness), mapping it to the Euclidean space $\mathbb{R}^n$ with $\log_{\mathbf{0}}^c$, performing matrix multiplication and mapping back to the hyperbolic space $\mathbb{D}_c^n$ with $exp_{\mathbf{0}}^c$. The only difference between the standard linear layer, in that case, is that log-exp treatment.

However, since we have a field of feature vectors, we could make use of the same approach and apply log-exp treatment to the convolution operation. Applying it to a feature vector field would mean performing exponential and logarithmic map, not over a single vector but every feature vector in an image. When comparing this to a case of Hyperbolic Linear Layer, see fig. 4.4, it arises as an almost natural extension of Hyperbolic Linear Layer to the case of convolutions.

Before proceeding to define hyperbolic convolution operation over feature gyrovector fields, it is necessary to define what applying exponential, logarithmic map and norm to a vector field means mathematically.

> **Definition 27: Riemannian Operations over Gyrovector Fields**
>
> Riemannian operations over gyrovector fields can be defined as operations over gyrovectors attached to every point of the field:
>
> $$\text{Exponential map: } \exp_{\mathbf{u}}^c(f) : \{f | f : \mathbb{R}^2 \to T_{\mathbf{u}}\mathbb{D}_c^n\} \to \{g | g : \mathbb{R}^2 \to \mathbb{D}_c^n\}$$
> $$f \mapsto \exp_{\mathbf{u}}^c(f) \quad \text{s. t.} \quad [\exp_{\mathbf{u}}^c(f)](\mathbf{x}) := \exp_{\mathbf{u}}^c(f(\mathbf{x}))$$
> $$\text{Logarithmic map: } \log_{\mathbf{u}}^c(f) : \{f | f : \mathbb{R}^2 \to \mathbb{D}_c^n\} \to \{g | g : \mathbb{R}^2 \to T_{\mathbf{u}}\mathbb{D}_c^n\}$$
> $$f \mapsto \log_{\mathbf{u}}^c(f) \quad \text{s. t.} \quad [\log_{\mathbf{u}}^c(f)](\mathbf{x}) := \log_{\mathbf{u}}^c(f(\mathbf{x}))$$
> $$\text{Norm: } \|f\| : \{f | f : \mathbb{R}^2 \to \mathbb{D}_c^n\} \to \{g | g : \mathbb{R}^2 \to \mathbb{R}\}$$
> $$f \mapsto \|f\| \quad \text{s. t.} \quad \|f\|(\mathbf{x}) := \|f(\mathbf{x})\|$$

Now, given the definition of Riemannian operations over fields of vectors, we can proceed to define convolution operation over fields of gyrovectors.

> **Definition 28: Möbius 2D Convolution**
>
> If conv2d $: \{f | f : \mathbb{R}^2 \to \mathbb{R}^{c_{in}}\} \to \{g | g : \mathbb{R}^2 \to \mathbb{R}^{c_{out}}\}$, is a Euclidean convolution operation, then $\forall \{f | f : \mathbb{R}^2 \to \mathbb{D}^{c_{in}}\}$:
>
> $$\text{conv2d}^{\otimes_c}(f) : \{f | f : \mathbb{R}^2 \to \mathbb{D}^{c_{in}}\} \to \{g | g : \mathbb{R}^2 \to \mathbb{D}^{c_{out}}\}$$
>
> $\text{conv2d}^{\otimes_c}(f) = \mathbf{0}$ \hfill if $\text{conv2d}(\log_0^c(f)) = \mathbf{0}$
>
> $\text{conv2d}^{\otimes_c}(f) = \exp_0^c(\text{conv2d}(\log_0^c(f)))$
>
> $\qquad = \dfrac{1}{\sqrt{c}} \tanh\left( \dfrac{\|\text{conv2d}(f)\|}{\|f\|} \tanh^{-1}(\sqrt{c}\|f\|) \right)$
>
> $\qquad \cdot \dfrac{\text{conv2d}(f)}{\|\text{conv2d}(f)\|}$ \hfill otherwise
>
> Note that we use norm as well as logarithmic and exponential maps defined over feature gyrovector field.
>
> An important implication of that is that we do not need to change the geometry of base space for suggested convolution to work, see fig. 4.5.

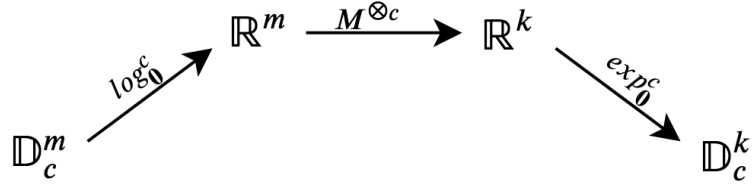

(a) Mappings between spaces in Hyperbolic Linear Layer.

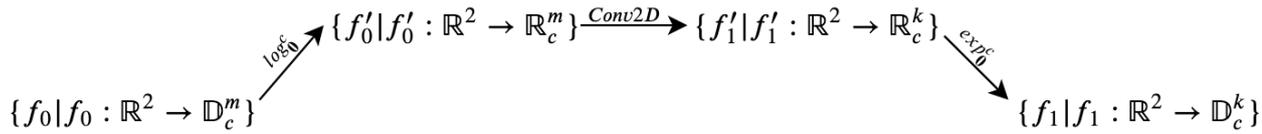

(b) Suggested mappings between spaces in Hyperbolic Convolutional Layer.

**Fig. 4.4.:** Mathematical view of mappings between spaces in Hyperbolic Linear and Convolutional layers.

> **Note**
>
> Here we have defined 2D convolution; however, without loss of generality, the same operation is applicable in the case of lower and higher-order convolutions with a change from *conv2d* to a corresponding operator.

There are two important points to notice about the suggested operations:

1. Since we perform the standard convolution operation, we can make use of all the low-level optimizations implemented by modern deep learning libraries.

2. We apply exponential, and logarithmic maps feature vector-wise; therefore, they allow for efficient parallelization on modern hardware.

So, after each Hyperbolic Convolutional Layer, we get an output as feature gyrovector field $f : \mathbb{R}^2 \to \mathbb{D}_c^{C_{out}}$, where $\mathbb{R}^2$ is a base manifold and $\mathbb{D}_c^{C_{out}}$ are feature gyrovectors attached to it.

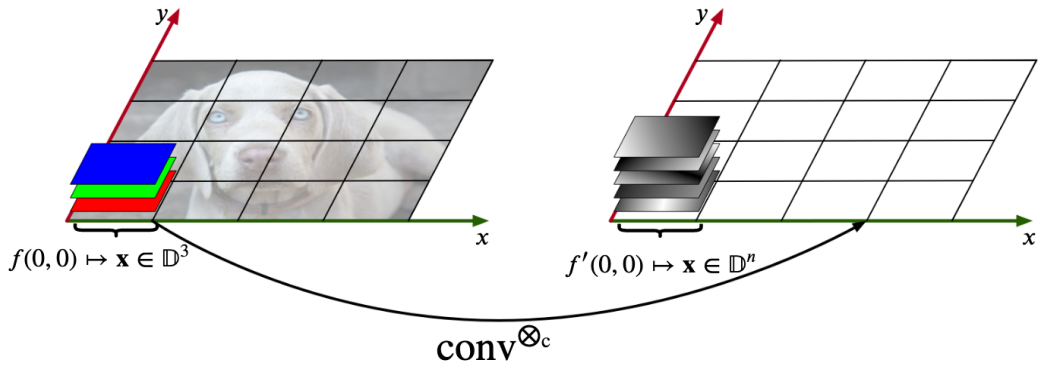

**Fig. 4.5.:** Visualization of images as feature gyrovector fields. Möbius Convolution operation is transforming feature gyrovectors attached to each point of a field.

Now, having defined a convolutional layer, we are ready to define operations that could be used to improve training.

### 4.4.2 Hyperbolic Batch Normalization

In section 4.1, we have discussed how batch normalization has been proven to be of use for training deep convolutional neural networks. Apart from speeding up the training, it is also leading model to better generalization, which is crucial for real-world applications of convolutional neural networks.

However, batch normalization algorithm 1 is only well-defined for the Euclidean space. To generalize it to the hyperbolic space, we need to redefine following operations and quantities: (1) Vector Addition; (2) Scalar Multiplication; (3) Mean; (4) Variance.

Generalization of the first two operations has already been discussed in section 2.4. We can directly apply gyrovector addition $\oplus_c$ and gyrovector scalar multiplication $\otimes_c$ instead of corresponding Euclidean operations.

However, calculating statistics over the batch is more problematic since we first need to generalize the definition of mean and variance to Riemannian manifolds. Generalization has been previously suggested by Maurice Frechet for metric spaces and then by Karsten Grove and Hermann Karcher for the general Riemannian manifolds (Grove and Karcher, 1973).

> **Definition 29: Frechet Mean and Variance**
>
> Let $(M, d)$ be a complete metric space. Let $x_i \in M, i \in \{1, \ldots n\}$ be points with corresponding weights $w_i \in \mathbb{R}, i \in \{1, \ldots n\}$. For point $p \in M$, **Frechet weighted variance** is defined as:
>
> $$\psi(p) = \sum_{i=1}^{N} w_i d^2(p, x_i)$$
>
> where $d$ is the distance function. The **Karcher means** $m \in M$ are then defined as points which locally minimize Frechet variance:
>
> $$m = \arg\min_{p \in M} \sum_{i=1}^{N} d^2(p, x_i), \quad (4.1)$$
>
> $$\text{where } d \text{ is the distance function.} \quad (4.2)$$
>
> If there is such a point $m \in M$ that globally optimizes $\psi$, it is called **Frechet mean**.

If $w_i = \frac{1}{n} \forall i \in \{1, \ldots n\}$, such Frechet variance is the direct generalization of standard variance to metric spaces.

Note that this definition is more general than that of the more familiar arithmetic mean. If $M \equiv \mathbb{R}$ and $d$ is the usual distance function, arithmetic mean is equal to a Frechet mean.

An apparent issue with calculating Frechet mean and variance, however, is that for most manifolds closed-form formula might not be available and can only be calculated as a result of a minimization problem. This yields prohibitively slow when applying in actual neural network architectures. Another issue is that such a point might be only locally optimal, which leads to a problem of choosing the optimal point.

However, for hyperbolic spaces, a unique, closed-form solution is provided by gyromidpoint, defined in (19). Calculation of such point is linear in the number of parameters provided, thus yielding performance which is on par with the Euclidean version of batch normalization in terms of efficiency. Given Frechet mean, we then can calculate Frechet variance with $w_i = \frac{1}{n} \forall i \in \{1,\ldots n\}$ as specified above with $d$ being the Möbius distance (see (18)).

These allow us to define Hyperbolic Batch Normalization.

---
**Algorithm 2** Hyperbolic Batch Normalization
---
**Input:** Value of $x$ over a mini-batch $\mathbb{B} = \{\mathbf{x}_1 \ldots \mathbf{x}_m\}, \mathbf{x}_i \in \mathbb{D}_c^n, i \in \{1, \ldots m\}$

**Input:** Parameters to be learned $\gamma \in \mathbb{R}, \boldsymbol{\beta} \in \mathbb{D}_c^n$

**Output:** $\{y_i = BN_{\gamma,\boldsymbol{\beta}}(\mathbf{x}_i)\}$

1: $\boldsymbol{\mu}_{\mathbb{B}} \leftarrow \frac{1}{2} \otimes_c \frac{\sum_{i=1}^m 2\gamma_{\mathbf{x}_i}^2 \mathbf{x}_i}{\sum_{i=1}^m (2\gamma_{\mathbf{x}_i}^2 - 1)}$ ▷ Mini-batch Mean

2: $\sigma_{\mathbb{B}}^2 \leftarrow \frac{2}{m\sqrt{c}} \sum_{i=1}^m \tanh^{-1}(\sqrt{c}\|-\mathbf{x}_i \oplus_c \boldsymbol{\mu}_{\mathbb{B}}\|)$ ▷ Mini-batch Variance

3: $\hat{\mathbf{x}}_i \leftarrow \frac{1}{\sqrt{\sigma_{\mathbb{B}}^2 + \epsilon}} \otimes_c (-\mathbf{x}_i \oplus_c \boldsymbol{\mu}_{\mathbb{B}})$ ▷ Normalize

4: $y_i \leftarrow \gamma \otimes_c \hat{\mathbf{x}}_i \oplus_c \boldsymbol{\beta} \equiv BN_{\gamma,\boldsymbol{\beta}}(\mathbf{x}_i)$ ▷ Scale and shift

---

Such definition of an algorithm is equivalent to its Euclidean version and, as we will show in the next chapters, leads to the improvements in the performance of hyperbolic neural networks.

### 4.4.3 Hyperbolic Pooling

Another important class of layers used in convolutional neural networks is pooling layers. They provide us with a way of downsampling feature maps, which results in models being able to extract more robust features as well as being more computationally efficient. Robustness comes from (a) the fact that the features extracted can be seen as a summarized version of the processed input features; (b) invariance to small deformations, which results in features being more robust and less position-dependent. Efficiency comes from the fact that as we apply more and more pooling layers, the dimensionality of the input of the convolutional layer decreases.

Two most common approaches to pooling are average pooling and max pooling. Both of them use the sliding window approach and apply the corresponding function[11] to each image patch. However, due to the difference in operations performed, they need to be generalized to the hyperbolic space separately.

---
[11] max for max-pooling or avg for average pooling.

Average pooling can be easily generalized to the hyperbolic space using previously defined Möbius midpoint (see (18)) with feature vectors corresponding to a given patch as inputs.

We can write it as: $f' = \frac{1}{2} \otimes_c \frac{\sum_{x=1,y=1}^{n} 2\gamma_{f(x,y)}^2 f(x,y)}{\sum_{x=1,y=1}^{n} (2\gamma_{f(x,y)}^2 - 1)}$, where $x, y \in \mathbb{R}$ are coordinates of a pixel that feature vector is attached to, $n$ is the height and width of the kernel.

Unfortunately, generalizing max pooling is not as straightforward because it breaks the hyperbolicity of the feature vectors in the output. To understand it, let us look at what max-pooling does at every step. While average pooling uses all the feature vectors as they are, max-pooling takes the largest component of input feature vectors in a channel-wise fashion.

What this results in is that the output feature vector might not be hyperbolic anymore. To solve this problem, we use exp-log-map approach. Therefore, before doing max pooling, we first project all feature vectors onto the tangent plane using a logarithmic map. Next, we apply max pooling, which results in a downsampled version of the input. Finally, we apply the exponential map to map from the tangent plane back to the manifold. Described sequence of operations results in the hyperbolic version of the classical max pooling.

Mathematically, it can be written as: $\text{maxpool}(f) = \exp_0^c(\max(\log_0^c(f)))$.

Such exp-log-map approach allows us to generalize any operation to the hyperbolic space, which we use for other layers that break hyperbolicity of the feature vectors, with dropout being the most critical example.

Definitions provided in this section are already enough to make most of the current convolutional neural networks architectures entirely hyperbolic. However, we can come up with a different approach in the case of graph convolutional networks, and this will be the primary matter of the next section.

## 4.5 Hyperbolic Graph Convolutional Networks

As mentioned in the previous chapter, another example of structured data except for images are graphs. In the last couple of years, there has been high interest in the Graph Neural Networks, which led to significant breakthroughs in graph processing models. The problem, however, was that all of the models assumed feature vector space of graphs to be Euclidean.

Despite being the most intuitive, Euclidean space might not be able to capture the intrinsic properties of the graph. This issue can be solved by using a more suitable embedding space (Weber and Nickel, 2018), which we expect to lead to better performance of the existing algorithms. As well as in the case of Hyperbolic Convolutional Neural Networks, this can be seen as a case of inductive bias. Despite being a beautiful idea theoretically, to account for those properties, a new graph neural network-based model is required. In this section, we tackle this task by making the GCNConv layer suggested in (Kipf and Welling, 2016) entirely hyperbolic.

Since the GCNConv layer consists of two matrix multiplications and one nonlinearity, we could use previously defined Möbius matrix multiplication and Möbius nonlinearity to make GCNConv layer hyperbolic. However, the main problem with that is that this would not affect graph neural networks performance since all operations except for the bias translation would be inherently Euclidean (fig. 3.5b). Such limitation, we conjecture, would prevent GCNConv layers from using hyperbolic space to its fullest. However, seeing the GCNConv layer from a message passing point of view allows us to derive an elegant and efficient way of embedding graphs in the hyperbolic space.

Since the message passing scheme for graphs consists of two steps - namely message passing and aggregation, we need to define hyperbolic versions of each one of those separately.

During message passing step, we multiply features of each node by a shared weight matrix and then pass these processed features to the adjacent nodes afterwards. We treat this operation as a standard hyperbolic linear layer defined in (Ganea et al., 2018a), which results in following definition of messages between neighbouring nodes $\mathbf{h}_v$ and $\mathbf{h}_w$: $M(\mathbf{h}_v, \mathbf{h}_w, \mathbf{W}) = \mathbf{W}^{\otimes_c} \mathbf{h}_w$. Afterwards, we pass these messages to the neighbouring nodes.

Similarly to the original message passing algorithm, we could use summation to aggregate incoming messages at each node. However, this would not be possible given that messages are hyperbolic feature vectors, since from the definition of the gyrogroup (see (12)) it follows that the operation of gyrovector addition is not associative and, therefore, will be dependent on the order of elements in the summation. Since operations in graphs have to be permutation-invariant, using summation as an aggregation function is highly undesirable.

The solution for this problem can be devised using ideas from Graph Attention Network (Veličković et al., 2017) and Hyperbolic Attention (Gulcehre et al., 2018) papers. Since we can treat normalization coefficients in GCNConv Layer similarly to attention coefficients in Hyperbolic Attention, we can use previously defined weighted
Möbius midpoint (see (20)) as an aggregation function. Therefore, given aggregation coefficients $c_{vw}$, we arrive to aggregated vector $m_v$ for node $v \in \mathcal{V}$ defined as
$m_v = \frac{1}{2} \otimes_c \frac{\sum_{w \in N(v)} 2c_{vw} \gamma^2_{M(\mathbf{h}_v, \mathbf{h}_w)} M(\mathbf{h}_v, \mathbf{h}_w)}{\sum_{w \in N(v)} (2c_{vw} \gamma^2_{M(\mathbf{h}_v, \mathbf{h}_w)} - 1)}$.

Using midpoint instead of sum, however, has a problem of having one less degree of freedom due to the normalization (see fig. 4.6). To add it back to the model, we multiple feature vector that we get after aggregation step by a scalar $\alpha$, which is, similarly to the weight matrix, common for all nodes in the graph.

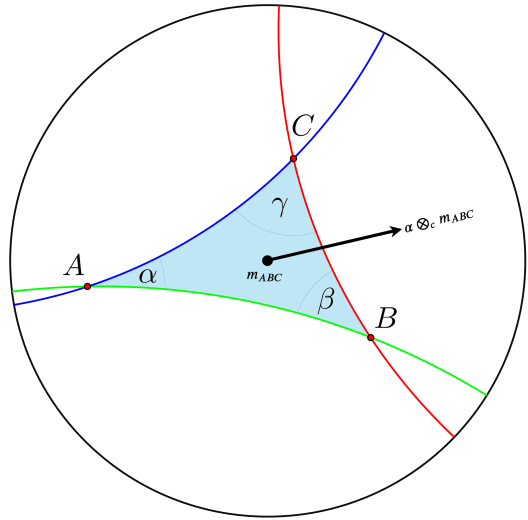

**Fig. 4.6.:** Midpoint loses one degree of freedom and can only lie within highlighted triangle. Multiplication by a scalar allows it to regain this lost degree of freedom; reproduced from Wikipedia.

Details mentioned above allows us to define HypGCNConv layer in its final form.

> **Definition 30: Hyperbolic GCNConv Layer**
>
> Given graph $\mathcal{G} = (\mathcal{V}, \mathcal{E})$, $\mathbf{h}_w \in \mathbb{D}_c^n$ is the feature vector of vertex $w$, $W \in \mathbb{R}^{m \times n}$, $\alpha \in \mathbb{R}$, $\deg(v)$ is the degree of the vertex $v$, $A$ is the adjacency matrix, $f : \mathbb{R}^m \to \mathbb{R}^m$ is the pointwise nonlinearity, HypGCNConv is defined in message passing terms as:
>
> $$M(\mathbf{h}_v, \mathbf{h}_w, \mathbf{W}) = \mathbf{W}^{\otimes_c} \mathbf{h}_w$$
> $$c_{vw} = (\deg(v) \deg(w))^{-\frac{1}{2}} A_{vw}$$
> $$m_v = \frac{\alpha}{2} \otimes_c \frac{\sum_{w \in N(v)} 2c_{vw} \gamma^2_{M(\mathbf{h}_v, \mathbf{h}_w)} M(\mathbf{h}_v, \mathbf{h}_w)}{\sum_{w \in N(v)} (2c_{vw} \gamma^2_{M(\mathbf{h}_v, \mathbf{h}_w)} - 1)}$$
> $$\mathbf{h}'_v = U_v(\mathbf{h}_v, m_v)$$
> $$= f^{\otimes_c}(m_v)$$
>
> More compactly, it can be written as:
>
> $$\mathbf{h}'_v = f^{\otimes_c} \left( \frac{\alpha}{2} \otimes_c \frac{\sum_{w \in N(v)} 2c_{vw} \gamma^2_{W^{\otimes_c} \mathbf{h}_w} W^{\otimes_c} \mathbf{h}_w}{\sum_{w \in N(v)} (2c_{vw} \gamma^2_{W^{\otimes_c} \mathbf{h}_w} - 1)} \right)$$

As we will see later, such implementation of the HypGCNConv layer allows us to make it as computationally efficient as its non-hyperbolic version while also making use of hyperbolicity of the space.

To give an outline of everything we did in this chapter, we would like to summarize all the operations we have defined so far in a single table.

These operations together with Hyperbolic Linear Layer and Hyperbolic Nonlinearity (as defined in (24), (25) and (23)) allow us to implement most modern architectures of deep convolutional neural networks. Details of the implementations will be the main matter of the next two chapters.

| Layer | Formula | Additional comments |
|---|---|---|
| Convolutional | $\text{conv2d}^{\otimes_c}(f) = \exp_0^c(\text{conv2d}(\log_0^c(f)))$ | exp and log are performed feature vector-wise; $f : \mathbb{R}^2 \to \mathbb{D}^{c_{in}}$ is an input image/stack of feature maps |
| Batch Normalization | see algorithm 2 | |
| Max Pooling | $\exp_0^c(\max(\log_0^c(f)))$ | max is performed channel-wise, exp and log are performed feature vector-wise; $f : \mathbb{R}^2 \to \mathbb{D}^{c_{in}}$ is the currently considered patch of an image/stack of feature maps |
| Average Pooling | $f' = \frac{1}{2} \otimes_c \frac{\sum_{x=1,y=1}^{n} 2\gamma_{f(x,y)}^2 f(x,y)}{\sum_{x=1,y=1}^{n} (2\gamma_{f(x,y)}^2 - 1)}$ | $f : \mathbb{R}^2 \to \mathbb{D}^{c_{in}}$ is the currently considered patch of an image/stack of feature maps |
| Dropout | $\exp_0^c(\text{dropout}(\log_0^c(f)))$ | exp and log are performed feature vector-wise; $f : \mathbb{R}^2 \to \mathbb{D}^{c_{in}}$ is an input image/stack of feature maps |
| Graph Convolutional | $\mathbf{h}'_v = f^{\otimes_c}\left( \frac{\alpha}{2} \otimes_c \frac{\sum_{w \in N(v)} 2c_{vw} \gamma_{\mathbf{W}^{\otimes_c} \mathbf{h}_w}^2 \mathbf{W}^{\otimes_c} \mathbf{h}_w}{\sum_{w \in N(v)} (2c_{vw} \gamma_{\mathbf{W}^{\otimes_c} \mathbf{h}_w}^2 - 1)} \right)$ | $\mathbf{h}_w \in \mathbb{D}_c^n$ is the feature vector of vertex $w$, $W \in \mathbb{R}^{m \times n}$, $\alpha \in \mathbb{R}$, $\deg(v)$ is the degree of the vertex $v$, $A$ is the adjacency matrix, $f : \mathbb{R}^m \to \mathbb{R}^m$ is the pointwise nonlinearity |

**Tab. 4.1.:** Set of operations necessary to implement hyperbolic versions of modern CNNs and GCNs.

# 5 Experimental Setup

In this chapter, we describe the experimental setup we will use to test models suggested in the previous section empirically. We start with describing the hyperbolic version of VGG model (Simonyan and Zisserman, 2014), which we call HypVGG. Then, we introduce a hyperbolic version of the GCN model described in (Kipf and Welling, 2016), which we call HypGCN. Next, we discuss architectural solutions we have used to improve the efficiency of the suggested models. Afterwards, we provide precise details of the experiments we will use to test our research questions outlined in section 1.1.

## 5.1 Models

We want to start by discussing the VGG model. Suggested by Simonyan and Zisserman (2014), VGG model was one of the first models contributing to significant breakthroughs in the field of Deep Learning. Due to its simplicity in terms of implementation yet excellent performance, it has been a go-to model since then allowing many researchers to validate their ideas. We choose to work with it due to the straightforward correspondence between hyperbolic and non-hyperbolic blocks, which simplifies the experimental setup.

There are multiple versions of VGG model, depending on the number of weight layers in the network. Throughout our experiments, we will use VGG11, which, in its hyperbolic version, we call HypVGG11. The original architecture of VGG models did not include batch normalization operation; however, since it led to improved performance, modern implementations of VGG add it after every Conv2D layer.

We define the hyperbolic version of VGG11 by making all the computational blocks hyperbolic. Moreover, we replace last linear layer with HypMLR layer, defined in (5).

As for GCN, we use the same setup as in (Kipf and Welling, 2016). Our model is a two-layer network with each layer being a GCNConv Layer. Such architecture allows us for a simple change of GCNConv Layers to HypGCNConv layers, which is sufficient to make model entirely hyperbolic. Optionally, similarly to HypVGG, we use HypMLR on top of the last HypGCNConv Layer. However, in this case, we also add nonlinearity and dropout before HypMLR layer.

Architectures of Euclidean versions of all models discussed above can be seen in fig. 5.1. Note that for visualization purposes, we split VGG11 architecture into feature extractor and classifier.

## 5.2 Additional Considerations

Although the straightforward, naive implementation of HypVGG model would be excessively expensive in terms of computations. The main reason for that is that the most straightforward implementation would have an unnecessary amount of exp and log maps between most layers.

This can be significantly simplified if we note that multiple sequential operations in $\mathbb{R}^n$ can be performed without back-and-forth mapping to $\mathbb{D}_c^n$.

Another critical point is that some of the hyperbolic operations need operands to already be in the hyperbolic space. Such statement is true, for example, for bias in case of bias addition layer. If it is the case that operand is not in the hyperbolic space when fed into a layer, we first apply the exponential map to it to project it onto the hyperbolic space. Another solution would be to use Riemannian optimization methods (Bécigneul and Ganea, 2018) instead. However, similarly to Mathieu et al. (2019), we argue that in the small learning rate regime, our solution and Riemannian optimization methods are equivalent.

## 5.3 Datasets

We use two different sets of data - images for VGG and signals on graphs for GCN.

As for images, we decided to use CIFAR10 (Krizhevsky et al., 2009) due to its simplicity and availability of benchmarks. CIFAR10 is a common dataset for testing modern convolutional neural networks in an image classification setting. It consists of $60,000$ RGB, $32 \times 32$, images split equally into 10 classes. $50,000$ images are used for training, while $10,000$ are used for testing.

We use common preprocessing for CIFAR10:

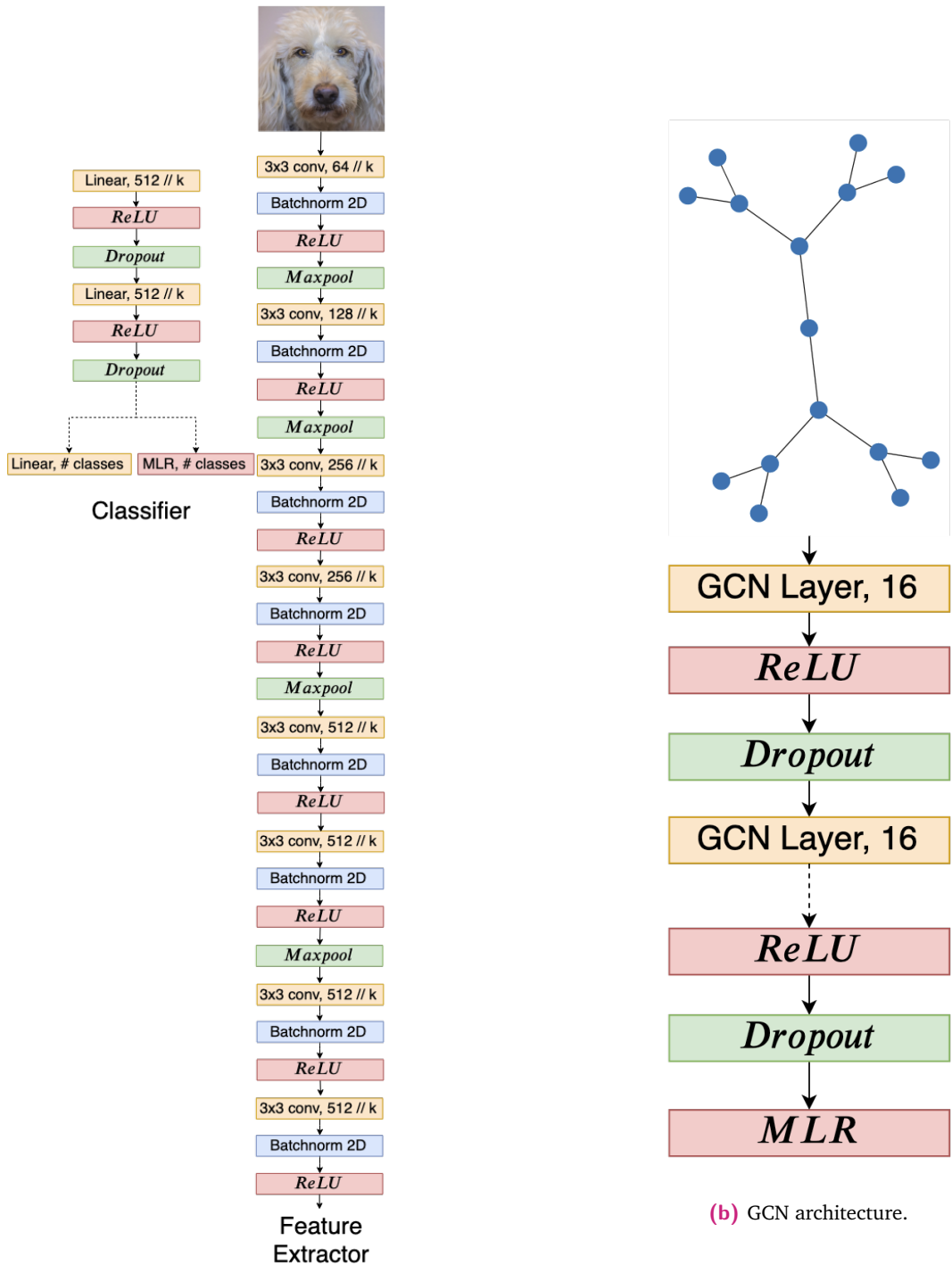

(a) VGG11 architecture. $k$ is varied as power of 2 from $2^0$ to $2^5$.

(b) GCN architecture.

**Fig. 5.1.:** VGG11 and GCN architectures.

- Normalization with $\mu = (0.485, 0.456, 0.406)$ and $\sigma = (0.229, 0.224, 0.225)$
- Random Horizontal Flip (only during training)
- Random Crop (only during training)

For graphs, we follow nearly the same setup as Kipf and Welling (2016). However, we only use citation networks in our experiments, namely Cora, Pubmed, and Citeseer (Sen et al., 2008). Graphs for citation networks are structured such that vertices are documents and edges are citations. One can see a summary of all three datasets in table 5.1.

| Dataset | Nodes | Edges | Classes | Features | Label rate |
| --- | --- | --- | --- | --- | --- |
| Citeseer | 3,327 | 4,732 | 6 | 3,703 | 0.036 |
| Cora | 2,708 | 5,429 | 7 | 1,433 | 0.052 |
| Pubmed | 19,717 | 44,338 | 3 | 500 | 0.003 |

**Tab. 5.1.:** Graph dataset statistics (Yang et al., 2016)

Feature vectors for each document are created using sparse bag-of-words model. Importantly, document nodes have class labels, and citation links are treated as undirected edges thus making it possible to use GCN for node classification. During training, all feature vectors are used, but only 20 nodes of each class are labelled. This approach casts problem to a domain of semi-supervised learning.

### 5.3.1 TreeDepth Dataset

Additionally to that, we create a proof-of-concept graph dataset, called TreeDepth, in large inspired by Mathieu et al. (2019). The primary motivation for creating a synthetic dataset is that hyperbolic spaces are well suited for the tree-like graphs, and by creating a synthetic graph dataset, we can ensure that graph being used is a tree. The algorithm is described in more details in appendix B; however, here, we will provide a high-level explanation of the dataset generation process.

We start by generating an initial vector $\mathbf{y_0} \in \mathbb{R}^n$ in the origin with depth $d = 0$. Next, we use it as a mean for a Normal distribution with unit variance. We sample $b$ vectors from this distribution, where $b$ is the branching factor which we set to $b = 2$. These $b$ nodes are now considered to be children of the initial vector. Next, we use sampled nodes as a means for new Normal distributions with unit variance. We then sample $b$ vector from each of these distributions. We repeat this process until desired depth $d$ is reached, which, in our case, is set to $d = 16$. After sampling all the vectors, we centre and normalize resulting dataset. Resulting vectors are treated as feature vectors of corresponding nodes. Depth of each node is chosen to be its label.

To make this dataset into a graph, undirected edges are drawn between parent and children nodes. Also, we choose feature dimensionality to be $n = 50$. Moreover, we split the dataset into train, validation, and test splits. Train set consists of 968 nodes, while validation consists of 531 nodes. All the other nodes are only observed during the test.

Summary of TreeDepth dataset can be found in table 5.2. Visualization of its simplified version is shown in fig. 5.2. More details on splits are provided in appendix B.

| Dataset   | Nodes  | Edges  | Classes | Features | Label rate |
|-----------|--------|--------|---------|----------|------------|
| TreeDepth | 131071 | 131070 | 17      | 50       | 0.0073853  |

**Tab. 5.2.:** TreeDepth dataset statistics.

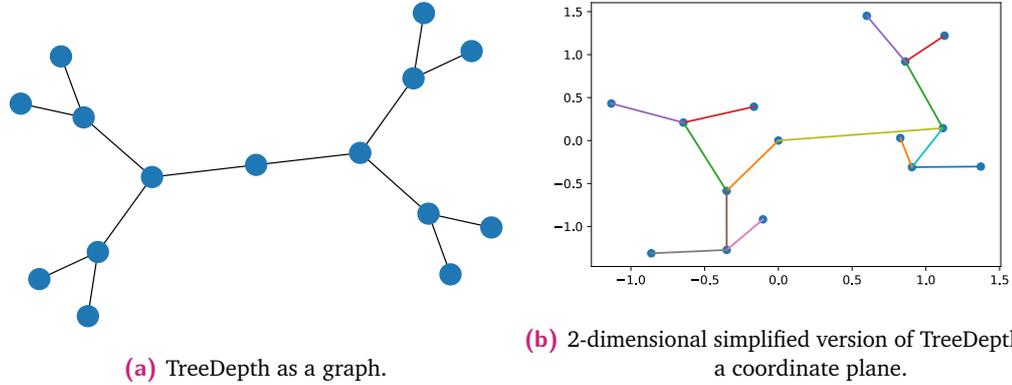

**(a)** TreeDepth as a graph.

**(b)** 2-dimensional simplified version of TreeDepth on a coordinate plane.

**Fig. 5.2.:** Visualization of simplified version of TreeDepth as a graph (left) and on a coordinate plane (right).

## 5.4 Experiments

We conduct two series of experiments: image classification and node classification.

Note that although we optimize hyperparameters for each experiment separately, we use Adam optimizer (Kingma and Ba, 2014) for all the experiments unless specified otherwise. Moreover, all the models are implemented using PyTorch deep learning library (Paszke et al., 2017).

### 5.4.1 Image Classification

This part is solely dedicated to investigating the performance of suggested blocks for CNNs. In this section, unless specified otherwise, we run experiments with a learning rate of $5e-3$, dropout set to 0, weight decay equal to $1e-3$, orthogonal initialization (Xiao et al., 2018) for Conv2D Layers and a batch size of 128.

First, we check whether adding BatchNorm layer to our network makes it learn faster. We start by checking whether HypVGG11 converges at all. We do so by running HypVGG11 for 50 epochs with a learning rate of $5e-4$. We expect that the model will be able to learn and reach accuracy far above such of a random guess.

5.4 Experiments | 65

Next, we compare HypVGG11 with and without BatchNorm layer by running both for 200 epochs with a learning rate of $5e-3$. We expect that BatchNorm layer would allow for a higher learning rate, thus yielding excellent performance for the model having it. At the same time, we expect that model without BatchNorm would not be able to learn under such a high learning rate, and its accuracy would stay approximately the same as such of a random guess.

Next, we run HypVGG11 with Batch Normalization and its corresponding Euclidean model VGG11 with Batch Normalization. We then compare the performance of VGG11 and HypVGG11 with different widths of layers by diving the original width of the layer by certain $k$, as shown in fig. 5.1a. In our experiments we vary $k \in \{2^0, \ldots 2^5\}$. We do so to see whether HypVGG11 performs better or worse than VGG11 in the low dimensional case.

VGG11 is used with optimal parameters, taken from `https://github.com/chengyangfu/pytorch-vgg-cifar10`. Each experiment is run 5 times to collect statistics. Finally, we take the model with the best performance and insert HypLinear layer before the final HypSoftmax layer. We set output dimensionality of HypLinear layer to 2 and retrain these two last layers while keeping other layers the same. This provides us with a visualization of image embeddings, which we use for qualitative analysis.

### 5.4.2 Node Classification

After that, we proceed to graph experiments. Following the approach in (Kipf and Welling, 2016), we perform cross-validation to find the optimal set of hyperparameters for HypGCN on Cora dataset. Interestingly, we find that completely removing ReLU nonlinearities in-between HypGCNConv layers improves performance by a large margin. This is consistent with observations in (Ganea et al., 2018a) and is explained by that exponential and logarithmic map can be seen as nonlinearities themselves.

GCN is used with standard hyperparameters as specified in the original paper (Kipf and Welling, 2016). Additionally, all the models are run with the early stopping of 10 steps with validation loss being the metric of interest.

We then use these optimal hyperparameters to test the performance of our model on all three datasets. Afterwards, we vary the embedding space dimensionality from $2^1$ to $2^4$ and compare the performance of both models in this low-dimensional setting. To account for different dimensionality, we find optimal hyperparameters for each dimensionality different from $2^4$ separately. Again, we run experiments 100 times to collect statistics.

Finally, we find optimal parameters for TreeDepth dataset for both GCN and HypGCN models. Since we have not found a significant difference in optimal hyperparameters for different dimensionality of the embedding space, we only find those for the dimension of $2^4$. Afterwards, models are run with these hyperparameters and embedding space varying between $2^1$ and $2^4$. These experiments are run 100 times as well to collect statistics. Additionally, we do an ablation study in order to achieve a better understanding of the difference between GCN and HypGCN models. It will be discussed in more details in the next chapter.

Tables with optimal hyperparameters for this section is provided in appendix A.

# 6 Results and Discussion

In this chapter, we present and discuss the results of the experiments conducted. We start by discussing the performance of HypVGG11 model on CIFAR10. First, we test HypVGG11 with and without Batch Normalization to see whether our definition of BatchNorm yields any benefits for the training of such a model. Next, we proceed to the comparison of the accuracy obtained by HypVGG11 and VGG11 with an optimal set of hyperparameters. Afterwards, we do a qualitative analysis of the feature embeddings coming from the last layer of HypVGG11. We conclude the chapter with an analysis of GCN and HypGCN performance on citation graphs and TreeDepth dataset.

## 6.1 Image Classification

Results presented in fig. 6.1 clearly show that suggested HypVGG11 model can learn. Due to a low learning rate, however, it learns slowly, reaching a test accuracy of 81.34% after 50 epochs. To speed-up the training, we use hyperbolic batch normalization technique, defined in algorithm 2.

Results of running HypVGG11 with and without batch normalization layer and learning rate of $5e-3$ confirm our assumptions regarding learning dynamics of these two models, see fig. 6.2. Namely, we can see that under such a high learning rate, HypVGG11 without batch normalization is not able to converge while batch normalization alleviates this issue. Also, importantly, our model can reach a test accuracy of 88% already after 25 epochs. The observed result serves as empirical proof of the correctness of the suggested HypBatchNorm layer.

Next, we proceed to the evaluation of optimal HypVGG11 + BN and VGG11 + BN models on CIFAR10. Results are presented in fig. 6.3 and table 6.1. It is clear from the results that suggested HypVGG11 model is not able to outperform VGG11, at least when using the original width of the convolutional layers. Interestingly enough, as we decrease the width, the performance gap gets smaller with the hyperbolic model beating the baseline at a width of 2.

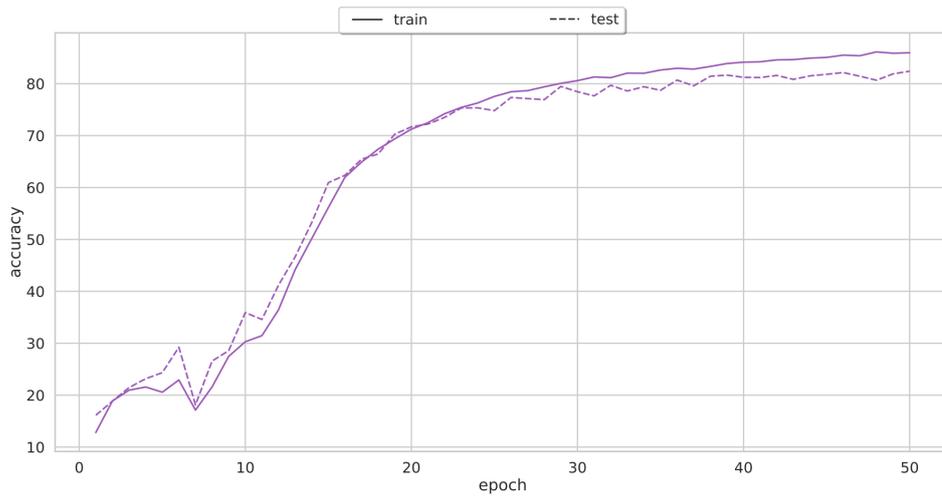

**(a)** Accuracy curve.

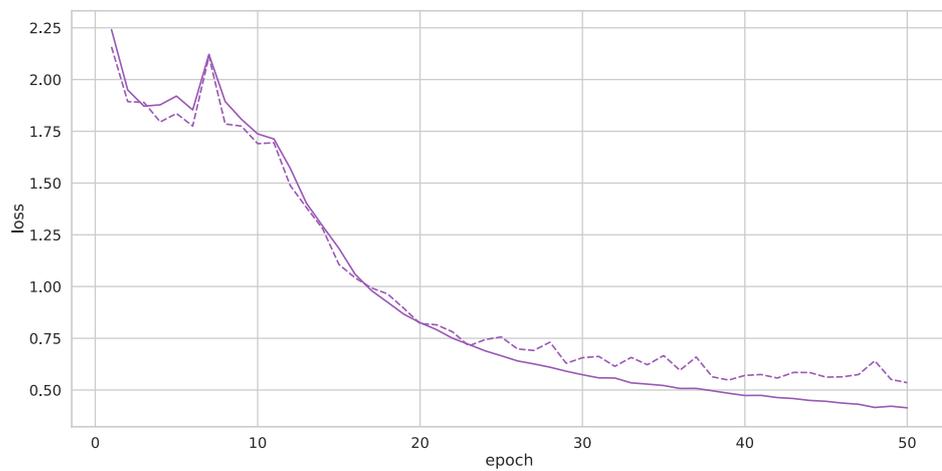

**(b)** Loss curve.

**Fig. 6.1.:** Performance of HypVGG11 model on CIFAR10 without batch normalization. Learning rate is set to $5e-4$.

Unsurprisingly, our results consistent with those presented in (Frogner et al., 2019). We use those together with the analysis of the dataset as an explanation for the observed performance.

The main reason for HypVGG11 performing worse is that CIFAR10 dataset is not suitable for the hyperbolic space. The fact that CIFAR10 has too few classes and, between themselves, classes do not form any hierarchy explains such behaviour. Since there are too few classes, our model is not able to extract hierarchical features from the images.

At the same time, the absence of hierarchy between datapoints themselves leads to the baseline outperforming our model, at least with large enough width. Since the width of the convolutional filter is the dimensionality of the hyperbolic space used for feature gyrovectors, lower-dimensional models can benefit from the higher capacity of the hyperbolic space compared to such of Euclidean space. This discussion is analogous to the one presented in section 3.1.2. One could argue that since CIFAR10 does not have graph representation, we do not have a way of measuring exact hyperbolicity of it. Such observation is correct; however, we hypothesize that, since CIFAR10 only has 10 classes, worst-case hyperbolicity would still be low enough so that our model works better in the lower-dimensional case.

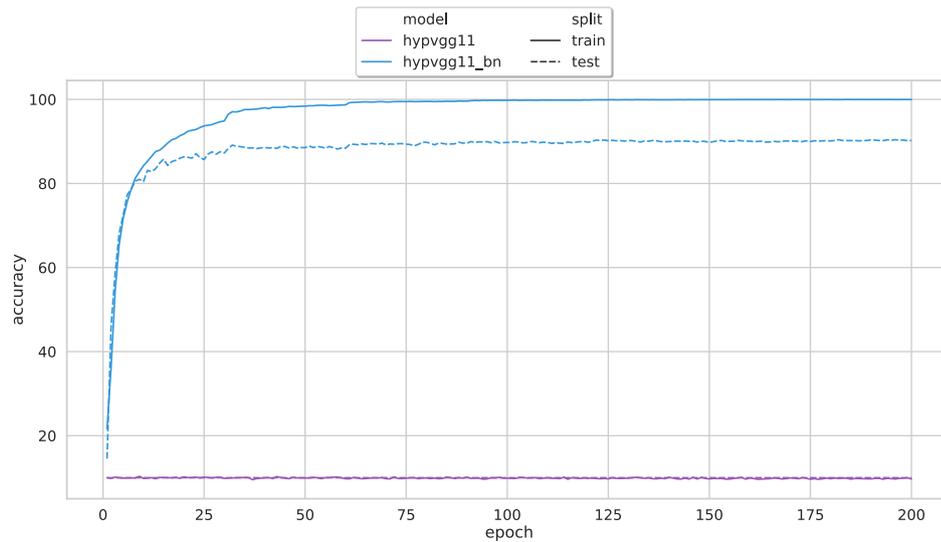

(a) Accuracy curve.

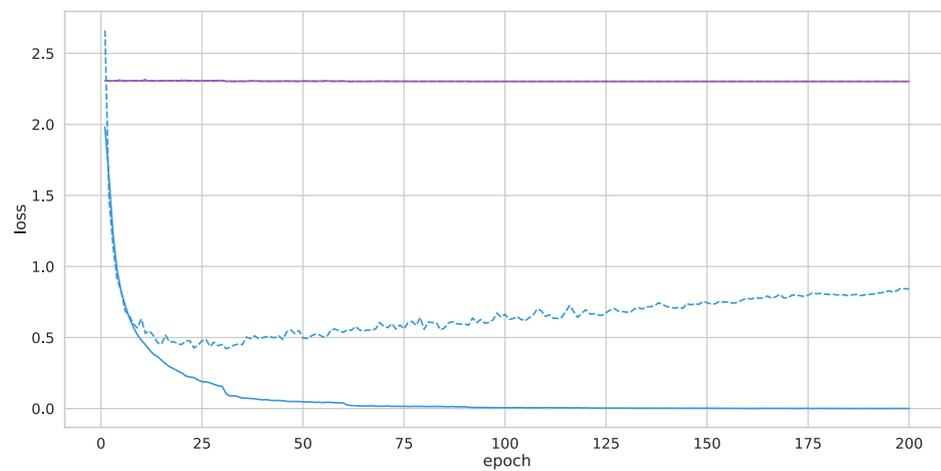

(b) Loss curve.

**Fig. 6.2.:** Performance of HypVGG11 model on CIFAR10 with and without batch normalization. Learning rate is set to $5e-3$.

Finally, to understand whether our model can learn meaningful image embeddings, we visualize 2-dimensional embedding of the output of the last layer. Results, provided in fig. 6.4, show that our model can successfully cluster most of the images from the same class together. At the same time, shapes of clusters indicate that the model can successfully use the power of hyperbolic softmax. As seen in fig. 3.4, hyperbolic softmax allows for more complex hyperplanes. Such behaviour stems from the fact that our data should now be linearly separable in hyperbolic space, which, due to the absence of parallel postulate provides us with more flexibility.

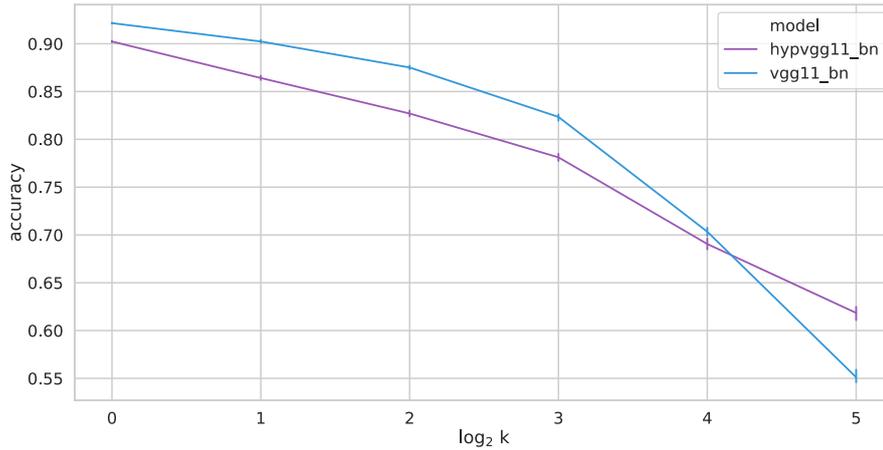

**Fig. 6.3.:** Accuracy of HypVGG11+BN and VGG11+BN on CIFAR10 dataset with different width $\frac{w}{k}$ of convolutional layers.

| k \ model | VGG11 | HypVGG11 | $\Delta$(gcn, hypgcn) |
|---|---|---|---|
| 1  | **92.17 ± 0.22781**  | 90.21 ± 0.42781 | 1.96 |
| 2  | **90.31 ± 0.75192**  | 86.49 ± 0.95192 | 3.82 |
| 4  | **87.55 ± 0.923591** | 82.78 ± 1.23591 | 4.77 |
| 8  | **82.34 ± 1.27204**  | 78.01 ± 1.67204 | 4.33 |
| 16 | **70.21 ± 1.856710** | 69.21 ± 2.56710 | 1 |
| 32 | 55.34 ± 2.60123      | **62.04 ± 3.160123** | 6.7 |

**Tab. 6.1.:** Performance of VGG11 and HypVGG11 models on CIFAR10 dataset. $k$ is the divisor of width. Highlighted is a better result.

Experiments presented in this subsection point to that suggested hyperbolic model is not able to outperform its Euclidean counterpart on CIFAR10. However, the margin between the two models is quite small, which can be potentially solved by finding more suitable hyperparameters.

This result, therefore, serves as a proof-of-concept for the usage of suggested HypCNN blocks, and further research is needed to understand whether worse performance comes from to the dataset or the model itself.

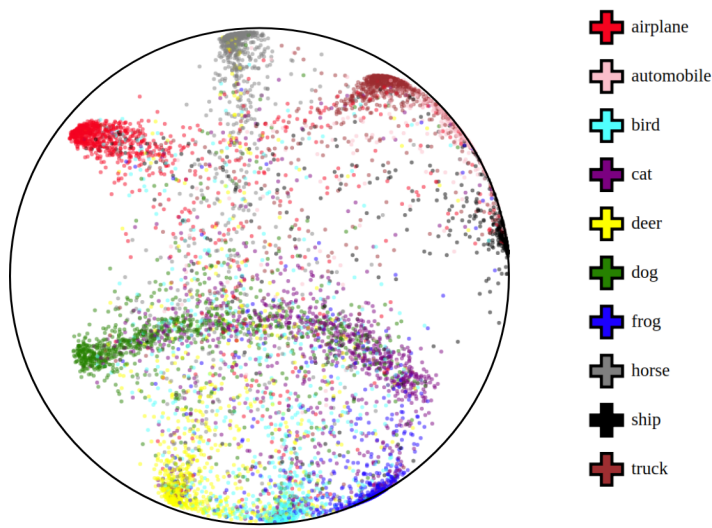

**Fig. 6.4.:** HypVGG11 image embedding.

## 6.2 Node Classification

We will now look at the performance of HypGCN compared to GCN on citation graphs. Results are presented in fig. 6.5. It is clear from the presented results that our model is underperforming compared to the baseline model by a large margin. Despite that the gap closes as the embedding dimension decreases, the baseline still performs better except for Citeseer dataset, where our model performs better in 2-dimensional case.

More detailed results can are presented in appendix C.

These results could be surprising if we would not have a framework of Gromov's hyperbolicity, allowing us to analyze graphs to understand whether the structure of the graph itself can explain the poor performance. However, with Cora and Citeseer, we instantly run into a problem of graphs not being connected, which, theoretically, prohibits us from measuring its hyperbolicity.

To alleviate this problem, we first find its connected components. Interestingly enough, the largest connected subgraph for both of these graphs accounts for most of the nodes. One can see detailed statistics of hyperbolicity of each graph in table 6.2, where "Largest" column contains information about the largest connected subcomponent. The final column shows hyperbolicity weighted by the size of the subcomponent divided by the total size of the graph.

These measurements explain why the margin between our model and the baseline is the smallest on Cora dataset and the largest on PubMed. We conjecture that the reason

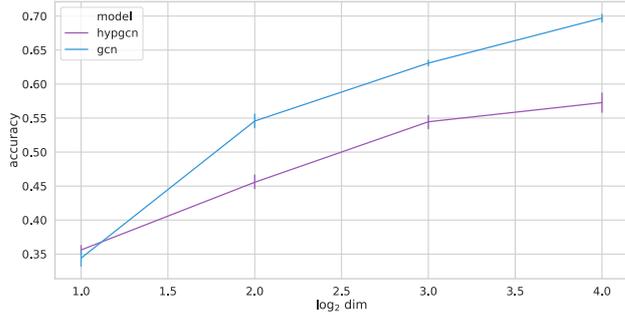

(a) Accuracy of GCN and HypGCN on Citeseer dataset with different dimensionality of the embedding space.

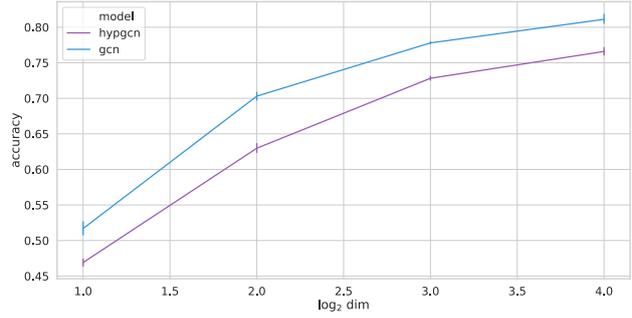

(b) Accuracy of GCN and HypGCN on Cora dataset with different dimensionality of the embedding space.

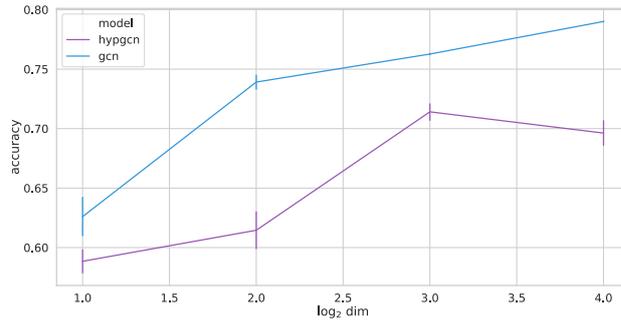

(c) Accuracy of GCN and HypGCN on PubMed dataset with different dimensionality of the embedding space.

**Fig. 6.5.:** Performance of HypGCN and GCN models on citation graphs.

| Dataset | Connected | Largest (# nodes) | $\delta$ of Largest | Weighted $\delta$ |
|---|---|---|---|---|
| **Citeseer** | No | 2120 (63.72 %) | 7.5 | 4.206041 |
| **Cora** | No | 2485 (91.76 %) | 4 | 3.689438 |
| **Pubmed** | Yes | 19717 (100 %) | 4.5 | 4.5 |

**Tab. 6.2.:** $\delta$-Hyperbolicity statistics for citation graphs.

for HypGCN outperforming GCN on Citeseer in 2-the dimensional case is due to a poor choice of hyperparameters for the Euclidean model. Such a conclusion is drawn based on the fact that we adopt an approach where we tune hyperparameters only on Cora instead of doing that on each dataset separately.

However, to confirm our hypothesis about that model performs poorly on graphs with high hyperbolicity, we would need to show the opposite behaviour on graphs with hyperbolicity of 0. That serves as the primary motivation for creating TreeDepth dataset.

Since TreeDepth is created in such a way that its underlying graph is a tree, it has hyperbolicity of 0. We expect it to result in HypGCN outperforming GCN by a large margin, which is confirmed by the results in fig. 6.6 and table C.4.

As expected based on the result of Frogner et al. (2019), as dimensionality increases, HypGCN outperforms GCN by a more considerable margin. Experimental results, therefore, serve as an empirical proof of the effectiveness of Hyperbolic Graph Neural Networks for the tree-like datasets. However, one might argue that the main increase in accuracy comes from using HypMLR since it can generate hyperplanes that are more suited for the tree structure, as shown in (Ganea et al., 2018a). To test it, we compare GCN and HypGCN architectures with both Euclidean and Hyperbolic MLR.

Results presented in table 6.3 clearly show that neither of the models is benefiting from having a version of softmax not natural to its original embedding space. GCN is not benefiting from having an additional linear + softmax layer on top at all and is thus omitted in the table.

However, even when having Euclidean MLR on top, HypGCN outperforms GCN. This result means that suggested HypGCN layers can successfully capture the tree structure of the graph while the same is not possible with standard GCN.

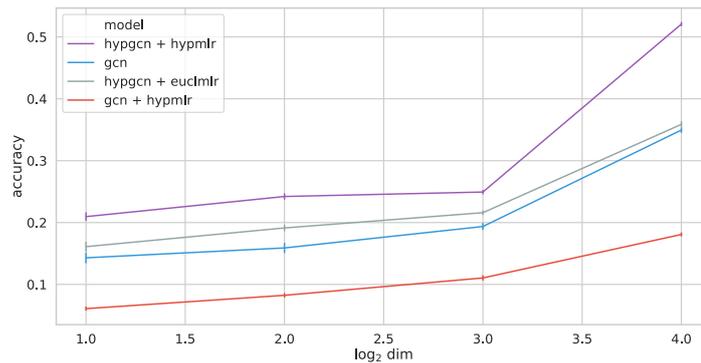

**Fig. 6.6.:** Accuracy of GCN and HypGCN on TreeDepth dataset with different dimensionality of the embedding space.

| Model | Accuracy |
|---|---|
| HypGCN + HypMLR | **52.23 % ± 3.5** |
| HypGCN + Euclidean MLR | 35.67 % ± 7.2 |
| GCN | 35.02 % ± 4.88 |
| GCN + HypMLR | 17.92 % ± 3.64 |

**Tab. 6.3.:** Performance of different models on TreeDepth dataset with $dim = 16$. Highlighted is the best result.

This result serves as a crucial empirical proof showing that, indeed, hyperbolic spaces are more suitable for tree-like datasets, and it is possible to use those in the context of deep neural models. However, it is still necessary to show its usefulness on the non-synthetic datasets, which we leave as a prospect for future work.

In this chapter, we have shown the results of running hyperbolic versions of standard image- and graph-processing models. Results show that models defined make sense and can succeed in those tasks. Nevertheless, to reach and, potentially, beat state-of-the-art results, suitable datasets are necessary, which we consider as one of the most critical points for future work.

# 7
# Conclusion

This work is aligned with research questions posed in section 1.1 and contributions mentioned in section 1.2. First, it introduces new computational blocks that enable modern neural network models to work with structured data in hyperbolic space. Secondly, it suggests a theoretical framework of Gromov hyperbolicity as an effective and computationally efficient tool for analyzing whether it is possible to embed data in the hyperbolic space with low distortion. Suggested framework proves itself effective by explaining results observed in recent literature. We also use it to explain the results observed in the experiments conducted.

Our experiments show that suggested models performance is comparable to the current state-of-the-art results. Although they do not outperform existing models on real-world datasets, we demonstrate that on a synthetic tree-like dataset, our model shows better results than the original non-hyperbolic model. This result suggests that the wrong selection of the datasets serves as an explanation for negative results. Moreover, it emphasizes the usefulness of Gromov hyperbolicity theory since it allows us to see if data can benefit from applying the hyperbolic version of the neural network before testing it empirically, which might be very computationally demanding.

Nevertheless, we believe that this work serves as an essential step towards bridging the gap between differential geometry and deep learning. To the best of our knowledge, suggested are the first convolutional architectures that are entirely hyperbolic and hope that these would spark an interest in the research community leading to more differential geometry-based deep learning.

## 7.1 Future Work

While working on this project, we have conceived many ideas that are worth further investigation and might lead to better machine learning models. We will discuss some of them in this section.

First of all, an exciting research direction is using more suitable datasets. There are already hierarchical datasets being used both for image classification (Deng et al., 2009) and graph embeddings (Miller, 1995). We believe that applying suggested models to those datasets should improve the performance of existing models by far margin. For ImageNet, suggested HypVGG can be used, while for WordNet, we can use R-GCN model (Schlichtkrull et al., 2018) with computational blocks changed to their hyperbolic versions [12] to make link prediction. To make it entirely hyperbolic, we can also use recently suggested scoring function (Balazevic et al., 2019) which embeds data in the hyperbolic space and by doing so, outperforms existing scoring function by a large margin. We have tried approaching both of these tasks in the process of working on this thesis; however, due to time constraints and the fact that R-GCN and ImageNet models are computationally expensive, we have not achieved satisfactory results to demonstrate in this work. Apart from the most well-known tree-like datasets, mentioned above, the medical and biological domain has a variety of tree-structured datasets (Fioravanti et al., 2018), which poses it as an exciting area for applying suggested models.

Secondly, suggested models are limited by the fact that we work in a tiny subspace[13], which leads to the need of using high-precision arithmetic and, thus, high computational costs and numerical instabilities. We think that it would be very beneficial if suggested operations could be generalized to the models of the hyperbolic space other than Poincaré disk, especially the Hyperboloid model. Despite being mathematically equivalent, the hyperboloid model has been shown to lead to better performance both in terms of speed and stability (Nickel and Kiela, 2018). Therefore, we expect that using such a model would be beneficial for models suggested in this work as well.

Another interesting observation that we made is that depending on the hyperbolicity, models using Euclidean space embeddings might beat Hyperbolic space embeddings at different dimensionality. If hyperbolicity is very high, Euclidean space performs better already at very low dimensions[14]. However, if it gets closer to zero, this "breakpoint" moves towards higher dimensions. Analyzing the relation between "breakpoint" dimensionality and hyperbolicity is a fascinating research question that might get us closer to using hyperbolic embeddings to their fullest.

Finally, an interesting idea would be to merge two models suggested in the paper by using robust graph embeddings for the benefit of the image classification. Since ImageNet has been build atop WordNet, there exists a one-to-one correspondence between classes in ImageNet and nodes in WordNet. This leads to a problem when trying to perform image classification, namely, that some classes are visually very similar. However, if we could embed data such that images are embedded close to their corresponding nodes, we could use it as an additional similarity function which should lead to better performance. A similar idea has been suggested by Barz and Denzler (2018). However, there authors use a sphere to embed WordNet tree, which, we hypothesize, is inherently worse than hyperbolic embeddings.

---

[12]This can be done similarly to how we changed GCN to HypGCN.
[13]$n$-sphere with $r = 1$.
[14]Starting from $dim = 2$ onwards.

# Websites

# List of Definitions

# List of Theorems

# List of Figures

# List of Tables

# A Hyperparameters for GCN and HypGCN models

| dataset | dimensionality | model | lr | dropout | weight decay | MLR | linear |
|---|---|---|---|---|---|---|---|
| Citation graphs | 16 | gcn | 0.01 | 0.5 | 0.0005 | False | False |
| | | hypgcn | 0.005 | 0.5 | 0.01 | True | True |
| | 8 | gcn | 0.05 | 0.2 | 0.001 | False | False |
| | | hypgcn | 0.01 | 0.2 | 0.001 | True | True |
| | 4 | gcn | 0.05 | 0 | 0.001 | False | False |
| | | hypgcn | 0.01 | 0.2 | 0.001 | True | True |
| | 2 | gcn | 0.05 | 0 | 0.001 | False | False |
| | | hypgcn | 0.05 | 0 | 0.001 | True | True |
| DeepTree | 16 | gcn | 0.1 | 0 | 0 | False | True |
| | | hypgcn | 0.01 | 0 | 0.0001 | True | True |

**Tab. A.1.:** Hyperparameters for used GCN and HypGCN models.

# Extra Details on TreeDepth Dataset

---

**Algorithm 3** Algorithm for generating TreeDepth Dataset
---
**Input:** Maximum depth $max\_d = 16$, branching $b = 2$, dimensionality $n = 50$, $\sigma_0 = 1$

**Output:** Dataset $out\_arr = \{y_i\}, \quad y_i \in \mathbb{R}^n$

 1: $y_0 \leftarrow \mathbf{0}$ ▷ Starting point
 2: $d \leftarrow 0$
 3: $out_arr \leftarrow \{y_0\}$ ▷ Initially array only contains starting point
 4: $cur\_arr \leftarrow \{y_0\}$ ▷ There is only the initial node at the depth 0
 5: **while** $d \leq max\_d$ **do** ▷ Repeat until maximum depth
 6:     $next\_arr \leftarrow \{\}$ ▷ No nodes for the next depth have been generated yet
 7:     **for** $y\_cur$ in $cur\_arr$ **do**
 8:         **for** $i \leftarrow 0$ to $b$ do **do** ▷ Each node at current depth generates $b$ branches
 9:             $y\_new \sim \mathcal{N}(y\_cur, \sigma_0)$ ▷ Generate new node from older one.
10:             ▷ $y\_cur$ is parent of $y\_new$.
11:             $next\_arr \leftarrow next\_arr + y\_new$ ▷ New node is 1 deeper.
12:         **end for**
13:     **end for**
14:     $cur\_arr \leftarrow next\_arr$ ▷ Update current nodes with their children
15:     $out\_arr \leftarrow out\_arr + next\_arr$ ▷ Append generated nodes to the final array
16: **end while**

---

| Depth | Total nodes | Train nodes | % Train of Total | Val nodes | % Val of Total | Test nodes | % Test of Total |
|---|---|---|---|---|---|---|---|
| 0 | 1 | 1 | 100 | 0 | 0 | 0 | 0 |
| 1 | 2 | 1 | 50 | 0 | 0 | 1 | 50 |
| 2 | 4 | 1 | 25 | 1 | 25 | 2 | 50 |
| 3 | 8 | 2 | 25 | 2 | 25 | 4 | 50 |
| 4 | 16 | 5 | 31.25 | 5 | 31.25 | 6 | 37.5 |
| 5 | 32 | 10 | 31.25 | 10 | 31.25 | 12 | 37.5 |
| 6 | 64 | 21 | 32.8125 | 21 | 32.8125 | 22 | 34.375 |
| 7 | 128 | 42 | 32.8125 | 42 | 32.8125 | 44 | 34.375 |
| 8 | 256 | 85 | 33.203125 | 50 | 19.53125 | 121 | 47.265625 |
| 9 | 512 | 100 | 19.53125 | 50 | 9.765625 | 362 | 70.703125 |
| 10 | 1024 | 100 | 9.765625 | 50 | 4.8828125 | 874 | 85.3515625 |
| 11 | 2048 | 100 | 4.8828125 | 50 | 2.44140625 | 1898 | 92.67578125 |
| 12 | 4096 | 100 | 2.44140625 | 50 | 1.220703125 | 3946 | 96.33789063 |
| 13 | 8192 | 100 | 1.220703125 | 50 | 0.6103515625 | 8042 | 98.16894531 |
| 14 | 16384 | 100 | 0.6103515625 | 50 | 0.3051757813 | 16234 | 99.08447266 |
| 15 | 32768 | 100 | 0.3051757813 | 50 | 0.1525878906 | 32618 | 99.54223633 |
| 16 | 65536 | 100 | 0.1525878906 | 50 | 0.07629394531 | 65386 | 99.77111816 |
| Total | 131071 | 968 | 0.7385310252 | 531 | 0.4051239405 | 129572 | 98.85634503 |

**Tab. B.1.:** Distribution of nodes in dataset splits per depth in TreeDepth dataset.

# C

# Extra GCN and HypGCN Results

| dataset | citeseer | | |
|---|---|---|---|
| dim \ model | **gcn** | **hypgcn** | \|Δ(gcn, hypgcn)\| |
| 2 | 34.4009 % ± 12.6226 | **35.6027 % ± 7.0603** | 3.7545 |
| 4 | **54.5661 % ± 10.4138** | 45.558 % ± 10.738 | 12.457 |
| 8 | **63.0893 % ± 4.6860** | 54.4580 % ± 10.0927 | 4.8571 |
| 16 | **70.3072 % ± 6.2690** | 57.2723 % ± 14.7434 | 9.3794 |

**Tab. C.1.:** Performance of GCN and HypGCN models on Citeseer dataset. Highlighted is a better result.

| dataset | cora | | |
|---|---|---|---|
| dim \ model | **gcn** | **hypgcn** | \|Δ(gcn, hypgcn)\| |
| 2 | **51.6920 % ± 9.7409** | 46.9107 % ± 6.0245 | 4.7813 |
| 4 | **70.3071 % ± 5.8712** | 62.9826 % ± 6.4378 | 7.3245 |
| 8 | **77.7963 % ± 1.9126** | 72.8225 % ± 3.0131 | 4.9738 |
| 16 | **81.5149 % ± 7.4950** | 76.5929 % ± 5.5281 | 4.522 |

**Tab. C.2.:** Performance of GCN and HypGCN models on Cora dataset. Highlighted is a better result.

| dataset | pubmed | | |
|---|---|---|---|
| dim \ model | **gcn** | **hypgcn** | \|Δ(gcn, hypgcn)\| |
| 2 | **62.6018 % ± 17.0248** | 58.8473 ± 10.5892 | 3.7545 |
| 4 | **73.9134 % ± 6.1287** | 61.4564 ± 16.8188 | 12.457 |
| 8 | **76.2679 % ± 0.6728** | 71.4108 ± 7.3089 | 4.8571 |
| 16 | **79.0035 % ± 0.4269** | 69.6241 ± 10.9589 | 9.3794 |

**Tab. C.3.:** Accuracy and standard error of GCN and HypGCN models on PubMed dataset. Highlighted is a better result.

| dataset | TreeDepth | | | |
| --- | --- | --- | --- | --- |
| model<br>dim | gcn | hypgcn + hypmlr | gcn + hypmlr | hypgcn + euclmlr |
| 2 | 14.9509 % ± 12.8295 | **21.4650 ± 9.3641** | 6.0885 % ± 3.7591 | 16.4390 % ± 10.2620 |
| 4 | 15.3727 % ± 10.7413 | **24.4473 ± 6.4631** | 8.2415 % ± 4.2587 | 19.1291 % ± 6.4390 |
| 8 | 19.9700 % ± 6.5267 | **25.0527 ± 4.1449** | 11.0486 % ± 5.1055 | 21.5958 % ± 4.1882 |
| 16 | 35.0209 % ± 4.8813 | **52.2255 ± 3.5013** | 17.923 % ± 3.64 | 35.6692 % ± 7.2003 |

**Tab. C.4.:** Accuracy and standard error of GCN and HypGCN models with different version of MLR on TreeDepth dataset. Highlighted is a better result.